\documentclass[journal]{IEEEtran}

\usepackage[utf8]{inputenc}
\usepackage[T1]{fontenc}
\usepackage{graphicx}
\usepackage{booktabs}
\usepackage{amsmath}
\usepackage{amssymb}
\usepackage{tikz}
\usetikzlibrary{arrows.meta,positioning,fit,calc}
\usepackage[hidelinks]{hyperref}
\hypersetup{
  pdftitle={A Benchmark and Layered Evaluation Protocol for Evidence-Grounded Shared Belief Updating in Connected Robotics},
  pdfauthor={Ranulfo Bezerra, Kimitoshi Yamazaki}
}

\graphicspath{{figures/}}

\begin{document}

\title{A Benchmark and Layered Evaluation Protocol for Evidence-Grounded
Shared Belief Updating in Connected Robotics}

\author{Ranulfo~Bezerra and~Kimitoshi~Yamazaki%
\thanks{R. Bezerra and K. Yamazaki are with Tohoku University, Sendai, Japan
(e-mail: ranulfo@tohoku.ac.jp; yamazaki@tohoku.ac.jp).}%
\thanks{This manuscript is under review.}%
\thanks{Project page: \url{https://creg-bench.github.io}}}

\markboth{Preprint under review}{Bezerra and Yamazaki: Evidence-Grounded Shared Belief Updating in Connected Robotics}

\maketitle
\bstctlcite{IEEEexample:BSTcontrol}

\begin{abstract}
Connected physical systems fuse observations from robots, IoT sensors, and edge-AI modules into a shared world belief, yet most evaluations score only the final answer or action and cannot tell whether a method attributed the evidence, updated its belief, handled contradictions, and justified its decision. We argue that such shared reasoning should be evaluated as an evidence-to-belief trajectory and introduce the Connected Robotics Evidence Graph (CREG) Benchmark to measure it. CREG comprises 350 deterministically generated scenarios across seven regimes (justified revision, belief preservation, relation association, contradiction onset and resolution, stale evidence, and source reliability), each with a normative posterior reference at every evidence prefix, verified by an independent log-odds checker and a blinded expert audit. Nine representation conditions separate relation content from serialization, ordering, and metadata, with residual confounds quantified. Evaluating four open-weight instruction models and six transparent fusion baselines yields three findings. First, entity grounding explains much of the weak-model deficit: weaker models lose trajectory score to outputs that never bind to the required entity, and in exploratory analysis their grounded updates approach those of stronger models. Second, deleting semantic relations costs 15.3 points of strict belief-state accuracy on relation-dependent scenarios (Holm-corrected $p<0.001$), yet an information-matched structured list outperforms the full graph and shuffling it does not hurt: relation content matters, graph serialization does not. Third, belief accuracy is strongly non-monotonic along the evidence sequence, with mid-trajectory failures that endpoint-only evaluation cannot detect. All scenarios, references, prompts, metrics, and code are available for offline, hardware-free reproduction at \url{https://creg-bench.github.io}.
\end{abstract}


\section{Introduction}
\label{sec:introduction}

Connected robotic systems combine observations from mobile robots, Internet of Things (IoT) sensors, fixed cameras, and edge-AI modules to monitor environments that no single agent observes well enough to act alone \cite{Bezerra2026ConnectedRobotics}. The scientific problem they pose is not connectivity or sensing in isolation but the construction of a shared, defensible belief about the physical world from evidence that is distributed, uncertain, delayed, and sometimes contradictory. For inspection, monitoring, and response tasks, the quality of that belief, not merely the action it triggers, determines whether the system can be trusted.

Existing evaluations rarely test this construction directly: most score a final answer, a task-success flag, or a selected action. Such endpoint measures cannot separate a system that reasoned correctly from one that reached the expected output by accident, for example by binding its prediction to the wrong physical entity, citing the wrong evidence, missing a contradiction, or revising too early or too late; conversely, a correct belief may support an evidence-supported action that differs from the single reference action. An end-task score collapses all of these cases into one number.

Connected Robotics makes this gap consequential. Reports carry source identity and reliability varies by source, so two agreeing reports are not two independent confirmations; evidence may arrive late, describe a state that has since changed, or be superseded by a correction; and semantic relations, such as which sensor measures which asset, determine whether an observation applies to the target entity at all. Each mechanism can yield a correct final answer for the wrong reason, or a defensible intermediate step that an endpoint metric scores as failure.

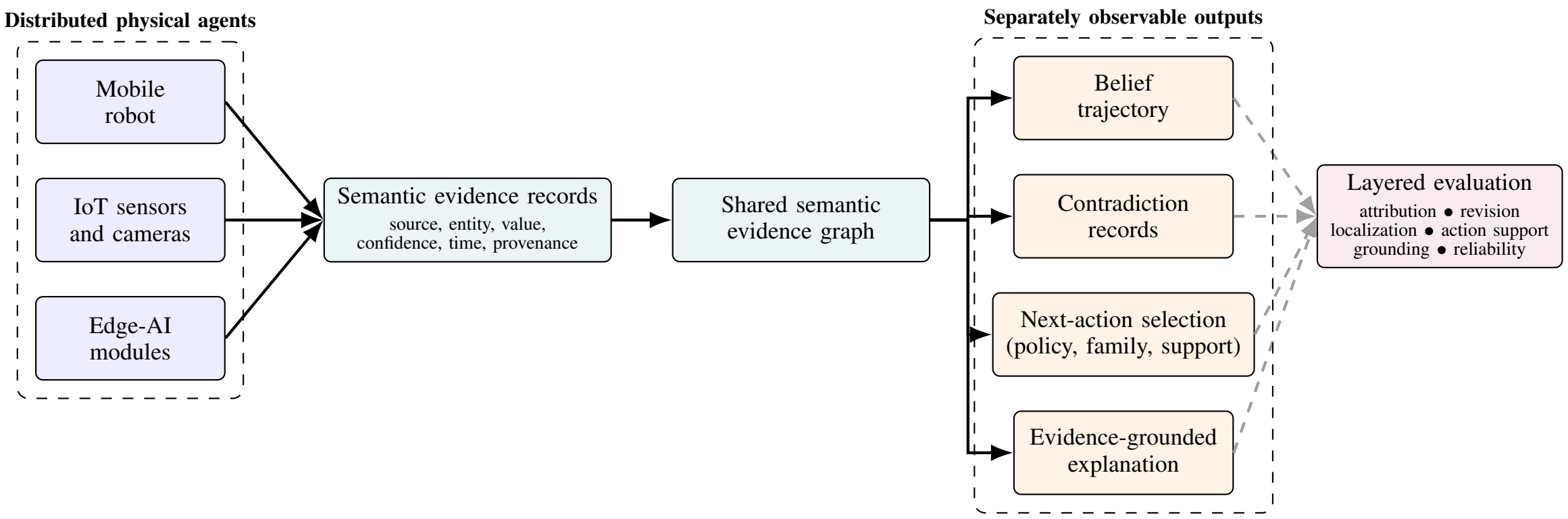
\begin{figure*}[!t]
    \centering
    \begin{tikzpicture}[
        scale=0.74,
        transform shape,
        font=\small,
        node distance=4.5mm and 8mm,
        box/.style={
            draw,
            rounded corners=2pt,
            align=center,
            minimum height=11mm,
            inner xsep=5pt,
            fill=white
        },
        source/.style={box, minimum width=25mm, fill=blue!7},
        stage/.style={box, minimum width=34mm, fill=teal!8},
        output/.style={box, minimum width=29mm, fill=orange!10},
        group/.style={
            draw,
            dashed,
            rounded corners=3pt,
            inner sep=5pt
        },
        flow/.style={-{Latex[length=2.2mm]}, thick},
        eval/.style={-{Latex[length=2mm]}, thick, dashed, gray!75}
    ]
        \node[source] (robot) {Mobile\\robot};
        \node[source, below=of robot] (iot) {IoT sensors\\and cameras};
        \node[source, below=of iot] (edge) {Edge-AI\\modules};

        \node[stage, right=13mm of iot] (records) {Semantic evidence records\\
            \scriptsize source, entity, value,\\[-1mm]
            \scriptsize confidence, time, provenance};
        \node[stage, right=of records] (graph) {Shared semantic\\evidence graph};

        \node[output, above right=5mm and 11mm of graph] (belief) {Belief\\trajectory};
        \node[output, below=of belief] (conflict) {Contradiction\\records};
        \node[output, below=of conflict] (action) {Next-action selection\\(policy, family, support)};
        \node[output, below=of action] (explanation) {Evidence-grounded\\explanation};

        \node[box, right=11mm of conflict, minimum width=32mm, fill=purple!8]
            (evaluation) {Layered evaluation\\
            \scriptsize attribution $\bullet$ revision\\[-1mm]
            \scriptsize localization $\bullet$ action support\\[-1mm]
            \scriptsize grounding $\bullet$ reliability};

        \draw[flow] (robot.east) -- (records.west);
        \draw[flow] (iot.east) -- (records.west);
        \draw[flow] (edge.east) -- (records.west);
        \draw[flow] (records) -- (graph);
        \draw[flow] (graph.east) -- ++(5mm,0) |- (belief.west);
        \draw[flow] (graph.east) -- ++(5mm,0) |- (conflict.west);
        \draw[flow] (graph.east) -- ++(5mm,0) |- (action.west);
        \draw[flow] (graph.east) -- ++(5mm,0) |- (explanation.west);
        \draw[eval] (belief.east) -- (evaluation.west);
        \draw[eval] (conflict.east) -- (evaluation.west);
        \draw[eval] (action.east) -- (evaluation.west);
        \draw[eval] (explanation.east) -- (evaluation.west);

        \node[group, fit=(robot)(iot)(edge),
            label={[font=\footnotesize\bfseries]above:Distributed physical agents}] {};
        \node[group, fit=(belief)(conflict)(action)(explanation),
            label={[font=\footnotesize\bfseries]above:Separately observable outputs}] {};
    \end{tikzpicture}

    \caption{Overview of CREG. Connected agents produce attributed semantic evidence organized in a shared graph; belief trajectory, contradiction records, next action, and explanation are exposed and scored separately, so an end-task success cannot conceal an earlier-layer error.}
    \label{fig:creg-overview}
\end{figure*}

We introduce the \emph{Connected Robotics Evidence Graph (CREG) Benchmark}, a controlled benchmark and layered evaluation protocol for observing the full evidence $\rightarrow$ belief $\rightarrow$ contradiction $\rightarrow$ action $\rightarrow$ explanation chain (Fig.~\ref{fig:creg-overview}). Scenarios are generated from a normative specification that fixes source reliability, temporal and supersession semantics, and relation-transfer rules, so that every evidence prefix has a reference posterior and the belief trajectory can be scored step by step. Representation conditions present the same facts as flat records, typed lists, or graphs and separate relation content from serialization, ordering, and metadata where possible, with the remaining prompt-length and joint-manipulation confounds quantified. Grounding, belief revision, contradiction handling, action selection (reference-policy compliance, action-family agreement, and evidence support), and explanation grounding are scored separately.

Our central thesis is that shared reasoning in Connected Robotics should be evaluated as an evidence-to-belief \emph{trajectory}, not as a final-answer problem. The contributions of this paper are:
\begin{enumerate}
    \item We formulate shared belief updating in Connected Robotics as a trajectory-level evaluation problem that separately measures entity grounding, evidence attribution, belief revision, contradiction handling, action selection (policy compliance, family agreement, and evidence support), and explanation grounding.
    \item We introduce CREG, a controlled benchmark of 350 distributed-evidence scenarios with per-prefix normative references, temporal and supersession semantics, source reliability, relation counterfactuals, and information-controlled representation conditions.
    \item We evaluate CREG across four open-weight instruction models and six transparent fusion baselines, showing that (i) entity grounding accounts for a substantial part of the weak-model deficit, (ii) semantic relation content improves categorical belief accuracy where relational evidence is required while graph serialization itself shows no consistent advantage, and (iii) final-prefix evaluation misses large transient failures along the belief trajectory.
\end{enumerate}
The CREG benchmark, normative references, prompts, configurations, evaluation code, and analysis scripts are available from the project page at \url{https://creg-bench.github.io}.

\section{Related Work}
\label{sec:related-work}

\subsection{Semantic World Models and Multi-Robot Scene Graphs}

3D scene graphs give robots structured abstractions above metric geometry: 3D Dynamic Scene Graphs, Kimera, and Hydra construct layered place--object--agent representations online, Hydra-Multi merges them across robot teams, and ConceptGraphs and HOV-SG open the vocabulary for perception, planning, and language-grounded navigation \cite{Rosinol2020DynamicSceneGraphs,Rosinol2021Kimera,Hughes2022Hydra,Chang2023HydraMulti,Gu2024ConceptGraphs,Werby2024HOVSG}. Closest to our setting, Semantic Belief Graphs and Semantic Belief Behavior Graphs couple uncertain semantic beliefs to navigation and inspection behaviors \cite{Ginting2023SemanticBeliefGraph,Ginting2024SemanticBeliefBehaviorGraph}, and heterogeneous inspection teams and the Connected Robotics formulation place such agents in a wider AI--IoT--robotics system \cite{Kojima2022HeterogeneousRobots,Bezerra2026ConnectedRobotics}. These works ask how to construct, merge, and act on world models. We ask a complementary question: given semantic observations with provenance, confidence, time, and possible conflict, does a reasoning method produce an appropriate shared belief, measured separately from whether its final action matches a policy? We propose no graph-construction pipeline and claim no intrinsic advantage for graph topology.

\subsection{Belief Fusion and Revision under Distributed Evidence}

Evidence-theoretic combination \cite{Dempster1967UpperLower}, covariance intersection under unknown cross-correlation \cite{Julier1997CovarianceIntersection}, and truth discovery under source dependence \cite{Dong2009SourceDependence,Li2016TruthDiscovery} motivate benchmark fields for confidence, provenance, and source identity, and scenarios in which redundant reports must not count as independent confirmation. Belief revision formalizes how a knowledge state should change under conflicting information, through consistency, incorporation, and minimal change in the AGM account and its iterated extension \cite{Alchourron1985TheoryChange,Darwiche1997IteratedBeliefRevision}. Recent language-model benchmarks operationalize parts of this problem: Belief-R tests retraction of defeasible conclusions and avoidance of unnecessary change \cite{Wilie2024BeliefRevision}, BeliefTrack scores update, preservation, and isolation over sequential evidence \cite{Xu2026BeliefTrack}, and BayesBench compares turn-wise model beliefs with Bayesian references \cite{Samanta2026BayesBench}. Their evidence is textual, logical, or simulated. CREG attaches reports to physical agents and entities, exposes time, confidence, provenance, and contradiction, and connects belief changes to operational actions; Table~\ref{tab:benchmark-comparison} summarizes the positioning against these and the embodied benchmarks below.

\begin{table*}[!t]
\centering
\footnotesize
\setlength{\tabcolsep}{3pt}
\caption{CREG versus the closest prior benchmarks ($\sim$: partial coverage; $\dagger$: concurrent preprint). The distinguishing feature is the joint observability of the full chain: attributed, time-stamped, possibly contradictory physical evidence in; per-prefix belief references and separately scored grounding, contradiction, action, and explanation layers out.}
\label{tab:benchmark-comparison}
\begin{tabular}{lcccccc}
\toprule
& Belief-R & BeliefTrack$^\dagger$ & BayesBench$^\dagger$ & ALFRED/TEACh/B-1K & OpenEQA & \textbf{CREG} \\
& \cite{Wilie2024BeliefRevision} & \cite{Xu2026BeliefTrack} & \cite{Samanta2026BayesBench} & \cite{Shridhar2020ALFRED,Padmakumar2022TEACh,Li2023Behavior1K} & \cite{Majumdar2024OpenEQA} & \textbf{(ours)} \\
\midrule
\multicolumn{7}{l}{\emph{Evidence setting}} \\
Sequential evidence with revision pressure & \checkmark & \checkmark & \checkmark & -- & -- & \checkmark \\
Reports attributed to distinct physical sources with reliability & -- & -- & -- & -- & -- & \checkmark \\
Per-report confidence, provenance, and timestamps & -- & -- & -- & -- & -- & \checkmark \\
Stale, delayed, and superseded evidence semantics & -- & -- & -- & -- & -- & \checkmark \\
Labeled contradictions (detection, localization, type) & -- & -- & -- & -- & -- & \checkmark \\
Relation-dependent cases with counterfactual pairs & -- & -- & -- & -- & -- & \checkmark \\
\midrule
\multicolumn{7}{l}{\emph{Evaluation targets}} \\
Probabilistic reference posterior at every evidence step & -- & $\sim$ & \checkmark & -- & -- & \checkmark \\
Belief preservation / unnecessary updates penalized & \checkmark & \checkmark & $\sim$ & -- & -- & \checkmark \\
Entity grounding scored separately from belief quality & -- & -- & -- & -- & -- & \checkmark \\
Evidence support scored separately from exact policy compliance & -- & -- & -- & $\sim$ & -- & \checkmark \\
Evidence-grounded explanations scored & -- & -- & -- & -- & -- & \checkmark \\
Information-controlled representation ablations & -- & -- & -- & -- & -- & \checkmark \\
\bottomrule
\end{tabular}

\end{table*}

\subsection{Neural and Graph-Conditioned Robotic Reasoning}

SayCan, PaLM-E, and RT-2 ground language-model decisions in learned affordances, embodied observations, and action tokens \cite{Ichter2023SayCan,Driess2023PaLME,Zitkovich2023RT2}; SayPlan plans over a hierarchical 3D scene graph with iterative correction \cite{Rana2023SayPlan}; and, outside robotics, Think-on-Graph and G-Retriever reason over or retrieve knowledge-graph structure \cite{Sun2024ThinkOnGraph,He2024GRetriever}. These systems motivate graph-conditioned evaluation, but they treat action success as the outcome and rarely control information content in graph-versus-text comparisons. A representation claim requires that graph and non-graph inputs preserve the same facts and differ only in controlled ways---serialization, ordering, identifiers, provenance, timestamps, and confidence---before an improvement can be assigned to structure rather than to prompt clarity or information loss. CREG approaches that requirement and quantifies where it falls short (Sec.~\ref{sec:methods-representations}).

\subsection{Embodied and Explanation Benchmarks}

ALFRED, TEACh, BEHAVIOR-1K, and OpenEQA define success at the instruction, dialogue, activity, or answer level \cite{Shridhar2020ALFRED,Padmakumar2022TEACh,Li2023Behavior1K,Majumdar2024OpenEQA}; none of these outcome measures determines whether distributed evidence was attributed, fused, and revised correctly. Rationale evaluation supplies a second precedent: ERASER separates agreement with annotated evidence from whether a rationale influenced the prediction \cite{DeYoung2020ERASER}, and faithfulness work distinguishes plausible explanations from faithful accounts of model behavior \cite{Jacovi2020Faithfulness}. We adopt the same decomposition: evidence-identifier overlap measures grounding coverage rather than faithfulness, detecting a contradiction is distinct from localizing it, and agreement with one deterministic policy measures compliance rather than decision quality. CREG brings these threads together: deterministic rules serve as a transparent reference policy, fusion strategies as inspectable comparison points, and language models are evaluated on the same frozen evidence under controlled representations.

\section{The CREG Benchmark}
\label{sec:methods}

\subsection{Scenario Suite and Normative References}
\label{sec:methods-suite}

The suite contains 350 scenarios generated deterministically from a normative specification. Each scenario concerns one target entity with a binary physical state (\texttt{anomalous} or \texttt{normal}) and a uniform prior. Evidence arrives as attributed reports, each carrying a source, a claimed value, a confidence, event and arrival times, and a state epoch. The specification fixes how these fields combine: a report's effective accuracy is $0.5 + (\mathrm{confidence} - 0.5)\cdot\mathrm{reliability}$, so source reliability modulates stated confidence; state epochs and explicit supersession links determine which reports still apply to the current state; and relation-transfer rules determine when a sensor's report reaches the target through a \texttt{measures} relation. Applying these rules to every cumulative evidence prefix yields a reference posterior, a categorical outcome (posteriors tied within $10^{-12}$ are \texttt{undetermined}), a probability direction, supporting- and counterevidence sets, and a contradiction state. Reference action rules define the normative next action at the endpoint.

Fifty scenarios are generated for each of seven regimes, each isolating one reasoning demand: \emph{justified revision} (applicable evidence must reverse the top belief), \emph{belief preservation} (weak counterevidence must not), \emph{relation association} (a \texttt{measures} relation is required to connect sensor evidence to the target), \emph{contradiction onset} (opposing same-epoch claims open an active contradiction), \emph{contradiction resolution} (a correction supersedes a conflicting claim), \emph{stale/delayed evidence} (older-epoch late arrivals are stale, whereas delayed current-epoch evidence applies), and \emph{source reliability} (reliability changes the posterior relative to confidence-only fusion). Relation-association scenarios are built as counterfactual pairs that become observationally indistinguishable once relations are deleted, so any performance difference on them is attributable to relation content. Across all scenarios the suite exposes 1{,}150 scenario--prefix pairs.

\begin{figure}[t]
\centering
\includegraphics[width=\linewidth]{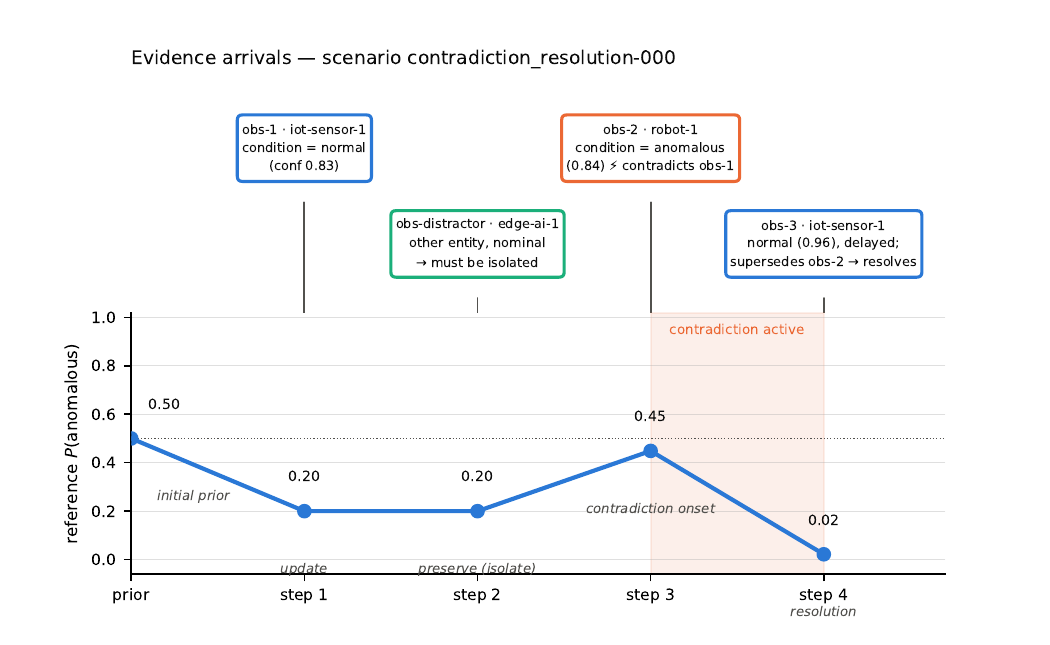}
\caption{One contradiction-resolution scenario. Top: four attributed arrivals; the robot's report contradicts the IoT sensor's, and a delayed high-confidence correction supersedes it. Bottom: the normative reference posterior scored at every prefix, with the contradiction-active interval shaded.}
\label{fig:worked-example}
\end{figure}

Fig.~\ref{fig:worked-example} traces one scenario: four arrivals from three heterogeneous sources exercise preservation, distractor isolation, contradiction onset, and supersession-based resolution, and the reference posterior at each prefix is the scored target. An independent checker that reconstructs every prefix posterior in log-odds space, without importing the generator's updater, agrees with the generator on all 1{,}150 prefixes.

\subsection{Expert Semantic-Plausibility Audit}
\label{sec:methods-audit}

Before any full model run, one independent expert reviewed a stratified sample of 35 scenario--prefix pairs (five per regime) under a blinded-then-reveal protocol: the reviewer recorded the expected top belief, probability direction, supporting and counterevidence, and contradiction status without access to the generated references, locked those answers, and only then compared them with the oracle. The first round accepted 29 of 35 cases; the six disagreements, concentrated in the relation-association, source-reliability, and stale/delayed regimes, traced to one specification issue (tied-posterior semantics), which was repaired at the source. The suite was regenerated and a second round accepted all 35 cases; no label was ever patched by hand. The audit is a semantic-plausibility check on the generator, not a measure of human consensus.

\subsection{Representation Conditions}
\label{sec:methods-representations}

\begin{figure*}[!t]
\centering
\begin{tikzpicture}[x=1cm, y=1cm, font=\footnotesize,
    ent/.style={draw=black!70, fill=black!5, rounded corners=2pt, inner sep=3pt,
        minimum height=5.5mm, font=\footnotesize\ttfamily},
    target/.style={ent, draw=black, thick},
    tag/.style={draw=black!50, fill=white, rounded corners=2pt, inner sep=3pt,
        align=left, font=\footnotesize},
    edge/.style={-{Latex[length=1.8mm]}, thick, black!75},
    elab/.style={font=\scriptsize\itshape, fill=white, inner sep=1pt},
    deleted/.style={-{Latex[length=1.8mm]}, thick, dashed, red!70!black},
    card/.style={draw=black!60, rounded corners=3pt, fill=white, inner sep=0pt},
    ctitle/.style={font=\footnotesize\bfseries, anchor=north west},
    csub/.style={font=\scriptsize\itshape, text=black!65, anchor=north west},
    mono/.style={font=\scriptsize\ttfamily, anchor=north west, align=left},
    band/.style={draw=black!40, rounded corners=3pt, fill=black!3},
    gnode/.style={draw=black!70, fill=black!5, rounded corners=2pt, inner sep=2.5pt,
        font=\scriptsize\ttfamily},
]
\node[band, minimum width=17.6cm, minimum height=1.6cm, anchor=north west] (band) at (0,0) {};
\node[anchor=north west, font=\footnotesize\bfseries] at (0.2,-0.08)
    {Semantic content of the source scenario, before representation manipulation};
\node[ent] (ag) at (2.0,-1.05) {iot-sensor-1};
\node[ent] (ob) at (5.0,-1.05) {obs-1};
\node[ent] (sa) at (7.9,-1.05) {sensor-a};
\node[target] (g1) at (10.9,-1.05) {gauge-1};
\draw[edge] (ag) -- node[elab, above] {reported} (ob);
\draw[edge] (ob) -- node[elab, above] {observes} (sa);
\draw[edge] (sa) -- node[elab, above] {measures} (g1);
\node[font=\scriptsize, text=black!65, below=0pt of g1] {scored target};
\node[tag, anchor=west] at (12.3,-1.0)
    {reading = anomalous\\ confidence = 0.91 \quad epoch = 1\\ source reliability = 0.91};
\foreach \x/\nm in {0/A, 4.5/B, 9.0/C, 13.5/D}
    \node[card, minimum width=4.1cm, minimum height=4.75cm, anchor=north west]
        (card\nm) at (\x,-1.95) {};
\node[ctitle] at (0.15,-2.05) {A. Canonical JSON};
\node[csub]   at (0.15,-2.45) {flat records + relation triples};
\node[mono]   at (0.15,-2.85) {%
    evidence: \{\\
    \ id: obs-1,\\
    \ source: iot-sensor-1,\\
    \ entity: sensor-a,\\
    \ value: anomalous,\\
    \ confidence: 0.91 \}\\[3pt]
    relation:\\
    \ \{sensor-a, measures,\\
    \ \ gauge-1\}};
\node[ctitle] at (4.65,-2.05) {B. Structured list};
\node[csub]   at (4.65,-2.45) {typed records; same facts};
\node[mono]   at (4.65,-2.85) {%
    [evidence]\\
    \ obs-1 | iot-sensor-1\\
    \ sensor-a | anomalous\\
    \ confidence 0.91\\[5pt]
    [relation]\\
    \ sensor-a | measures\\
    \ gauge-1};
\node[ctitle] at (9.15,-2.05) {C. Full graph};
\node[csub]   at (9.15,-2.45) {same facts as explicit edges};
\node[gnode] (c1) at (10.2,-3.1) {iot-sensor-1};
\node[gnode] (c2) at (10.2,-3.95) {obs-1};
\node[gnode] (c3) at (10.2,-4.8) {sensor-a};
\node[gnode, draw=black, thick] (c4) at (10.2,-5.65) {gauge-1};
\draw[edge] (c1) -- node[elab, right=3pt] {reported} (c2);
\draw[edge] (c2) -- node[elab, right=3pt] {observes} (c3);
\draw[edge] (c3) -- node[elab, right=3pt] {measures} (c4);
\node[font=\scriptsize\itshape, text=black!65, anchor=north west, text width=3.8cm]
    at (9.15,-6.0) {evidence on sensor-a transfers to \mbox{gauge-1}};
\node[ctitle] at (13.65,-2.05) {D. Graph without relations};
\node[csub]   at (13.65,-2.45) {semantic relation deleted};
\node[gnode] (d1) at (14.7,-3.1) {iot-sensor-1};
\node[gnode] (d2) at (14.7,-3.95) {obs-1};
\node[gnode] (d3) at (14.7,-4.8) {sensor-a};
\node[gnode, draw=black, thick] (d4) at (14.7,-5.65) {gauge-1};
\draw[edge] (d1) -- node[elab, right=3pt] {reported} (d2);
\draw[edge] (d2) -- node[elab, right=3pt] {observes} (d3);
\draw[deleted] (d3) -- node[midway, fill=white, inner sep=1pt, font=\small\bfseries,
    text=red!70!black] {$\times$} node[elab, right=8pt, text=red!70!black] {measures} (d4);
\node[font=\scriptsize\itshape, text=red!70!black, anchor=north west, text width=3.8cm]
    at (13.65,-6.0) {evidence on sensor-a no longer transfers to \mbox{gauge-1}};
\node[band, minimum width=17.6cm, anchor=north west, inner sep=4pt, font=\scriptsize]
    at (0,-6.95) {\textbf{Additional controlled factors (full graph, Qwen2.5-7B only):}\enspace{}%
    shuffled record order\enspace$\cdot$\enspace{}flat edge-removed relations\enspace$\cdot$\enspace{}%
    no provenance\enspace$\cdot$\enspace{}no timestamps\enspace$\cdot$\enspace{}no confidence};
\end{tikzpicture}
\caption{Information control in the representation study, on one observation of a frozen relation-association scenario. Top: the source scenario's attributed observation and \texttt{measures} relation. Conditions A--C preserve this content and differ only in serialization; D removes the \texttt{measures} relation, so relation-dependent evidence cannot be associated with the scored target. The footer lists the factor ablations; confounds are stated in Sec.~\ref{sec:methods-representations}.}
\label{fig:representations}
\end{figure*}
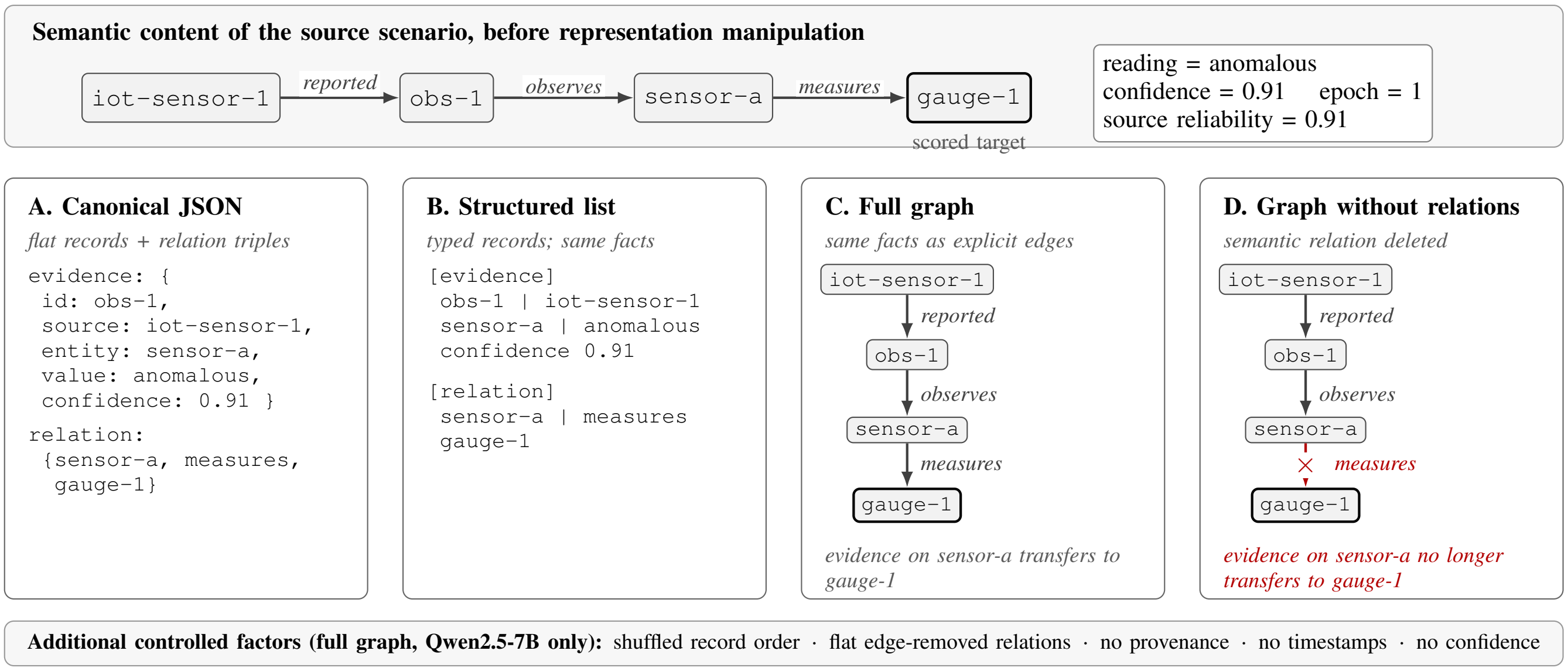

The representation study separates serialization changes from targeted deletion of semantic or metadata content (Fig.~\ref{fig:representations}). \emph{Canonical JSON} (flat records with relation triples), \emph{structured list} (typed records), and \emph{full graph} (reported, observes, and semantic-relation edges) preserve the same scenario information, whereas \emph{graph without relations} deliberately removes semantic-relation content; these four conditions are crossed with all models. Five factor ablations run on Qwen-7B only: \emph{flat edge-removed relations} (relation content kept, graph serialization removed), \emph{no provenance}, \emph{no timestamps}, \emph{no confidence}, and \emph{shuffled full graph} (identical records in permuted order). The shuffled graph and the flat edge-removed condition are length-matched controls (length ratios 1.000 and 1.004 to the full graph). Two manipulations carry confounds that constrain interpretation and are stated here once: canonical JSON prompts average $0.843\times$ the full-graph length, so serialization and prompt length covary; and the no-provenance condition also removes the source-reliability field that the normative likelihood consumes, so it is a joint manipulation. Remaining factor-level caveats are collected in Sec.~\ref{sec:discussion-limitations} and the appendix.

\section{Evaluation Setup}
\label{sec:evaluation}

\subsection{Model Panel and Runtime Protocol}
\label{sec:methods-panel}

The panel comprises Qwen2.5-1.5B-Instruct (Qwen-1.5B), Qwen2.5-7B-Instruct (Qwen-7B), Mistral-7B-Instruct-v0.3 (Mistral-7B), and Gemma-3-12B-IT (Gemma-12B), with fixed decoding (sampling, temperature 0.2, top-$p$ 0.9), three generation seeds treated as repeated generations nested within the same 350 scenarios, and a 512-token output budget selected by a runtime pilot. Under the preregistered compute fallback, every model runs canonical JSON and Qwen-7B alone runs the complete representation and factor-ablation matrix. Qwen-1.5B and Mistral-7B, which cannot reliably ground their outputs in the required entity, are retained as a \emph{diagnostic} tier so that their grounding failures are reported rather than filtered; Gemma runs on the three other principal representations are exploratory. Benchmark, references, prompts, metrics, and runtime configuration were frozen before analysis; selection details, budget sensitivity, and the freeze record are in the appendix.

\subsection{Evaluation Layers and Metrics}
\label{sec:methods-metrics}

Each output layer is scored separately. \emph{Reliability}: strict JSON validity and the rate of outputs recovered by offline repair; missing or malformed outputs score zero on categorical and F1 metrics and receive the worst-case trajectory distance. \emph{Grounding}: whether the prediction binds to the required physical entity, with failures split into missing-entity and hallucinated-entity rates. \emph{Belief trajectory}: the primary score is $1 - \overline{\mathrm{TV}}$, one minus the mean total-variation distance between predicted and reference posteriors over all reference prefixes; strict prefix belief-state accuracy, belief-preservation accuracy, correct-revision rate, Brier distance to the normative posterior (agreement with the reference, not calibration), and update latency complete the family. A \emph{conditional} variant that averages only over produced predictions is an exploratory, unpreregistered diagnostic; because it conditions on a non-random subset of prefixes it upper-bounds update quality. \emph{Evidence attribution}: supporting- and counterevidence F1 against the reference sets. \emph{Contradiction handling}: detection accuracy, localization F1 over the involved evidence identifiers (scored only when a contradiction is present), and type accuracy. \emph{Action selection} (endpoint task, canonical-JSON systems): exact reference-policy compliance, action-family agreement, and an evidence-supported-action rate that asks whether the chosen action is justified by the graph evidence irrespective of the reference action; the support predicate is a scenario-level function of the endpoint graph (active contradiction, maximum belief uncertainty, minimum report confidence) and on this suite accepts \texttt{verify} or \texttt{inspect} in 300 scenarios and only \texttt{monitor} in the 50 relation-association scenarios (Sec.~\ref{sec:results-baselines}). No separate action-admissibility metric is reported: the references annotate a single reference action per scenario. \emph{Explanation grounding}: coverage of the required evidence identifiers, i.e., citation coverage rather than faithfulness.

\subsection{Reference Systems and Transparent Baselines}
\label{sec:methods-baselines}

Six transparent fusion baselines run on the identical suite: latest-observation, unweighted majority, confidence-weighted, recency-weighted, source-reliability-weighted, and a Dirichlet-style updater. These provide fully inspectable reference strategies against which neural belief trajectories can be compared. The deterministic rule pipeline (graph construction, contradiction detection, confidence-sum fusion, rule-based action selection) is reported as a transparent reference implementation: its action row measures agreement between this operational policy and the normative reference policy, which are different policies by design, and its belief row is the measured accuracy of its fusion rule against the normative labels.

\subsection{Statistical Analysis}
\label{sec:methods-statistics}

The scenario is the independent unit; generation repeats are averaged within scenario before inference. Point estimates carry 95\% scenario-clustered bootstrap confidence intervals (10{,}000 resamples). Paired continuous contrasts use two-sided sign-flip permutation tests (100{,}000 resamples), paired binary outcomes use exact McNemar tests on the scenario-level fully-grounded indicator, and Holm correction is applied within each preregistered family (eleven representation contrasts, including three regime-scoped versions, and two model contrasts, per metric). The exploratory Gemma comparison receives confidence intervals only. Regime-level analyses and evidence-prefix curves never treat prefixes as independent scenarios. All contrasts, metrics, exclusion rules, and seeds were preregistered before the full run; the conditional trajectory score and the Gemma non-canonical systems are the two labeled exceptions.

\section{Results}
\label{sec:results}

Table~\ref{tab:main-results} reports the primary trajectory metrics for the panel and Table~\ref{tab:contrasts} the paired contrasts that support the three findings below, with scenario-clustered confidence intervals and Holm-adjusted permutation $p$-values. Unless stated otherwise, differences quoted in this section are Holm-significant at $p<0.01$; the complete contrast family is in the appendix.

\begin{table*}[!t]
\centering
\footnotesize
\setlength{\tabcolsep}{5pt}
\caption{Primary trajectory results (350 scenarios; three seeds averaged within scenario; 95\% scenario-clustered bootstrap CIs). Diagnostic-tier systems retain their grounding failures. Gemma non-canonical systems are exploratory (Appendix).}
\label{tab:main-results}
\begin{tabular}{lcccc}
\toprule
System & Trajectory score [95\% CI] & Strict state acc. & Grounding & Strict valid \\
\midrule
Qwen-1.5B / JSON (diagnostic) & 0.397 [0.380, 0.413] & 0.331 & 0.786 & 0.807 \\
Mistral-7B / JSON (diagnostic) & 0.239 [0.229, 0.248] & 0.221 & 0.694 & 0.897 \\
Qwen-7B / JSON & 0.797 [0.789, 0.806] & 0.831 & 1.000 & 1.000 \\
Gemma-12B / JSON & 0.801 [0.794, 0.808] & 0.729 & 1.000 & 0.999 \\
Qwen-7B / structured list & 0.842 [0.829, 0.854] & 0.682 & 0.966 & 0.999 \\
Qwen-7B / flat edge-removed & 0.774 [0.760, 0.788] & 0.605 & 0.917 & 0.999 \\
Qwen-7B / graph w/o relations & 0.727 [0.710, 0.743] & 0.537 & 0.868 & 1.000 \\
Qwen-7B / full graph & 0.814 [0.808, 0.821] & 0.695 & 1.000 & 1.000 \\
Qwen-7B / graph, no provenance & 0.831 [0.824, 0.838] & 0.730 & 1.000 & 1.000 \\
Qwen-7B / graph, no timestamps & 0.825 [0.818, 0.831] & 0.680 & 0.997 & 1.000 \\
Qwen-7B / graph, no confidence & 0.805 [0.798, 0.813] & 0.716 & 1.000 & 1.000 \\
Qwen-7B / shuffled graph & 0.848 [0.840, 0.856] & 0.741 & 0.999 & 1.000 \\
\bottomrule
\end{tabular}

\end{table*}

\begin{table*}[!t]
\centering
\footnotesize
\setlength{\tabcolsep}{5pt}
\caption{Paired contrasts supporting the main findings (Qwen-7B unless a model contrast is named). $\Delta$ is the mean paired scenario-level difference; regime-scoped rows restrict to the named regime; two-sided sign-flip permutation tests with Holm correction within each preregistered family. The Gemma contrast is exploratory (CI only).}
\label{tab:contrasts}
\begin{tabular}{llrrr}
\toprule
Contrast (A $-$ B) & Metric & $\Delta$ & 95\% CI & Holm $p$ \\
\midrule
\multicolumn{5}{l}{\emph{Relation content}} \\
Full graph $-$ graph w/o relations & traj. score & $+0.088$ & $[+0.072, +0.103]$ & $0.0001$ \\
\quad same, relation-association regime & traj. score & $+0.020$ & $[+0.002, +0.037]$ & $0.0577$ \\
\quad same, relation-association regime & strict acc. & $+0.153$ & $[+0.098, +0.207]$ & $0.0001$ \\
\multicolumn{5}{l}{\emph{Serialization and order}} \\
Full graph $-$ flat edge-removed & traj. score & $+0.041$ & $[+0.028, +0.053]$ & $0.0001$ \\
Structured list $-$ canonical JSON & traj. score & $+0.045$ & $[+0.033, +0.056]$ & $0.0001$ \\
Shuffled graph $-$ full graph & traj. score & $+0.034$ & $[+0.025, +0.042]$ & $0.0001$ \\
\multicolumn{5}{l}{\emph{Metadata fields (regime-scoped)}} \\
Full graph $-$ no provenance, source-reliability regime & strict acc. & $-0.200$ & $[-0.233, -0.167]$ & $0.0001$ \\
Full graph $-$ no timestamps, stale/delayed regime & strict acc. & $-0.003$ & $[-0.033, +0.027]$ & $0.9149$ \\
\multicolumn{5}{l}{\emph{Models (canonical JSON)}} \\
Qwen-7B $-$ Qwen-1.5B & traj. score & $+0.401$ & $[+0.382, +0.419]$ & $<0.0001$ \\
Qwen-7B $-$ Mistral-7B & traj. score & $+0.559$ & $[+0.545, +0.572]$ & $<0.0001$ \\
Gemma-12B $-$ Qwen-7B (exploratory) & traj. score & $+0.004$ & $[-0.005, +0.013]$ & -- \\
\bottomrule
\end{tabular}

\end{table*}

\subsection{Entity Grounding Accounts for Much of the Weak-Model Deficit}
\label{sec:results-grounding}

\begin{figure*}[!t]
\centering
\includegraphics[width=0.82\textwidth]{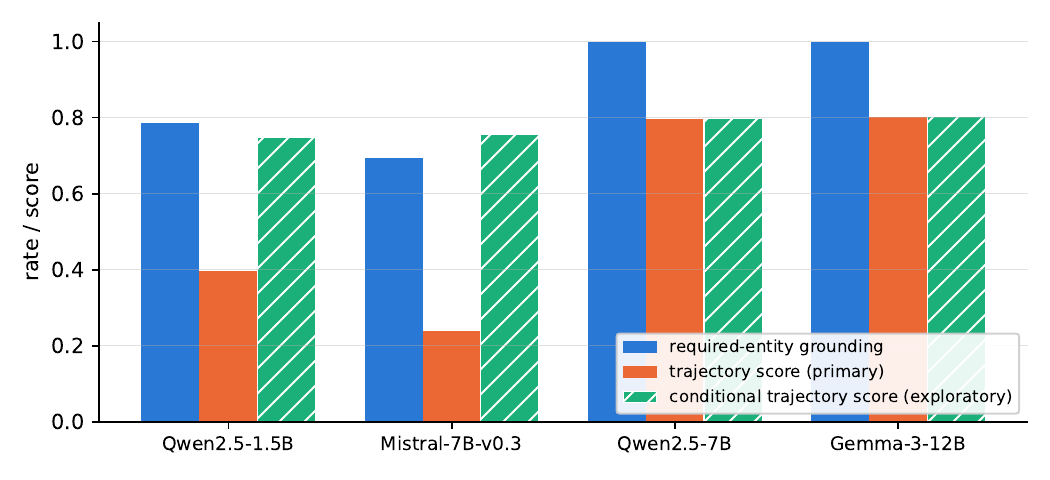}
\caption{Required-entity grounding rate, primary (unconditional) trajectory score, and exploratory conditional trajectory score (hatched; grounded predictions only, an upper bound) for the four models under canonical JSON.}
\label{fig:grounding}
\end{figure*}

Required-entity grounding under canonical JSON ranges from 0.694 (Mistral-7B) to 1.000 (Qwen-7B) across the four models (Fig.~\ref{fig:grounding}), and failures are almost entirely missing-entity outputs: hallucinated-entity rates are at most 0.016 for any system. The two diagnostic-tier models emit mostly valid JSON (strict validity 0.807 for Qwen-1.5B and 0.897 for Mistral-7B) yet reach unconditional trajectory scores of only 0.397 and 0.239. Qwen-7B exceeds Qwen-1.5B by 0.401 trajectory-score points ($[0.382, 0.419]$) and Mistral-7B by 0.559 ($[0.545, 0.572]$), with strict-accuracy gaps of 0.499 and 0.609, and exact McNemar tests on the scenario-level fully-grounded indicator separate both pairs ($p<10^{-98}$).

The exploratory conditional decomposition locates much of this deficit before belief updating. Averaged only over prefixes with a grounded prediction, the diagnostic models score 0.747 (Qwen-1.5B) and 0.755 (Mistral-7B), within a few points of the core models, whose conditional and unconditional scores nearly coincide (Qwen-7B: 0.797 in both; Gemma-12B: 0.802 versus 0.801). Grounding also varies along the evidence sequence (Fig.~\ref{fig:prefix-curves}b): Mistral's grounded-prediction rate is 0.009 at the first prefix, 0.051 at the second, 0.860 at the third, and 0.470 at the fourth, whereas Qwen-7B and Gemma-12B remain at or near 1.0 throughout. Weak models therefore do not merely produce worse posterior updates; they frequently fail before belief updating can be assessed, because their outputs never bind to the required physical entity. Entity grounding thus accounts for a substantial fraction of the observed weak-model deficit. Because the conditional score is unpreregistered and conditions on a non-random, plausibly easier subset of prefixes, it upper-bounds update quality, and we do not conclude that grounding is the only source of the gap.

In the exploratory model comparison, Gemma-12B matches Qwen-7B on trajectory score ($\Delta=+0.004$, $[-0.005, +0.013]$) but scores 0.101 lower on strict state accuracy ($[-0.116, -0.087]$); its predicted-versus-reference confidence-bin agreement is also lower (0.787 versus 0.905, Appendix).

\subsection{Semantic Relations Matter; Graph Serialization Does Not Consistently Help}
\label{sec:results-representations}

Relation content is necessary for categorical belief accuracy in the relation-association regime, whose counterfactual pairs are indistinguishable without relations: deleting semantic relations reduces Qwen-7B's strict belief-state accuracy there by 15.3 points ($[0.098, 0.207]$, Holm-adjusted $p<0.001$). The corresponding regime-scoped trajectory-score difference is smaller ($+0.020$, $[0.002, 0.037]$) and does not survive Holm correction ($p=0.058$). Separately, deleting relations reduces the trajectory score by 0.088 over all regimes ($[0.072, 0.103]$, Holm $p<0.001$). Relation content therefore matters where the task is constructed to require it, but not every relation-sensitive metric registers the effect at the regime level.

Graph serialization itself shows no consistent advantage. The information-matched structured list scores 0.842 against 0.814 for the full graph (Table~\ref{tab:main-results}), and the shuffled full graph---identical records in permuted order---scores 0.848, which is 0.034 above the ordered graph ($[0.025, 0.042]$). In the other direction, the flat edge-removed condition, which retains relation content without graph serialization, scores 0.041 below the full graph ($[0.028, 0.053]$), and canonical JSON trails the graph by 0.017. Representing the same facts as graph edges thus helps relative to one flat encoding, hurts relative to another, and record order carries no measurable value. We accordingly attribute no consistent advantage to graph serialization or to record order.

The two primary metrics rank serializations differently. Canonical JSON trails the graph on trajectory score ($-0.017$) but exceeds it on strict belief-state accuracy by 0.136 ($[0.120, 0.151]$), and exceeds the structured list on strict accuracy by 0.148 while trailing it on trajectory score by 0.045. Because canonical prompts are $0.843\times$ the graph's length for identical information, whereas the shuffled and edge-removed controls are length-matched, the canonical comparison confounds serialization with prompt length and is not assigned to format alone.

\begin{figure*}[!t]
\centering
\includegraphics[width=0.8\textwidth]{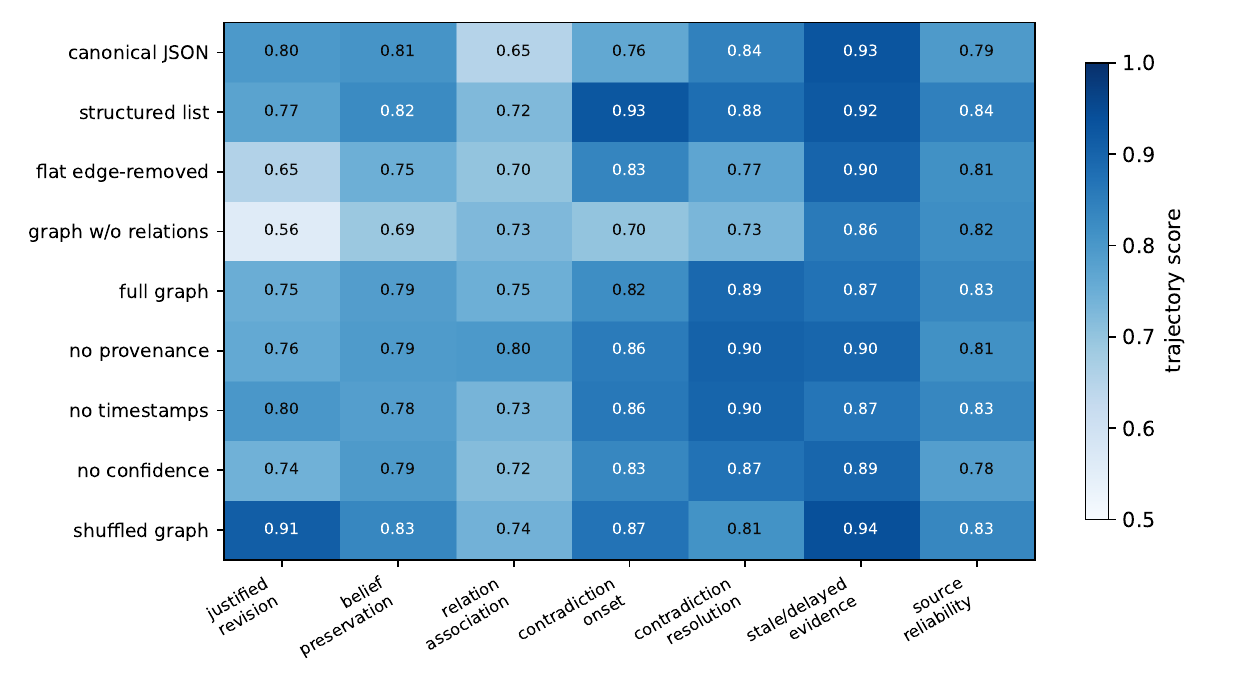}
\caption{Trajectory score by representation condition and regime (Qwen-7B, nine conditions, common 0.5--1.0 scale). Effects vary by regime; the strong relation-specific categorical effect is reported in Table~\ref{tab:contrasts}, and the regime-scoped trajectory-score effect is small.}
\label{fig:heatmap}
\end{figure*}

Among the field ablations (Fig.~\ref{fig:heatmap}), removing provenance changes the overall trajectory score by $+0.017$ in favor of the ablated condition, the source-reliability-regime trajectory score by 0.019 in favor of the full graph, and strict accuracy on that regime by 0.200 in favor of the ablated condition. Removing timestamps yields a null result on the stale/delayed regime (strict accuracy $\Delta=-0.003$, $[-0.033, 0.027]$, $p=0.91$) and $-0.010$ overall in favor of the ablated condition; removing confidence costs 0.009. These small, direction-inconsistent effects do not support a clean per-field story (Sec.~\ref{sec:discussion-representation}).

\subsection{Trajectory Evaluation Reveals Failures Hidden at the Endpoint}
\label{sec:results-trajectories}

\begin{figure*}[!t]
\centering
\includegraphics[width=0.88\textwidth]{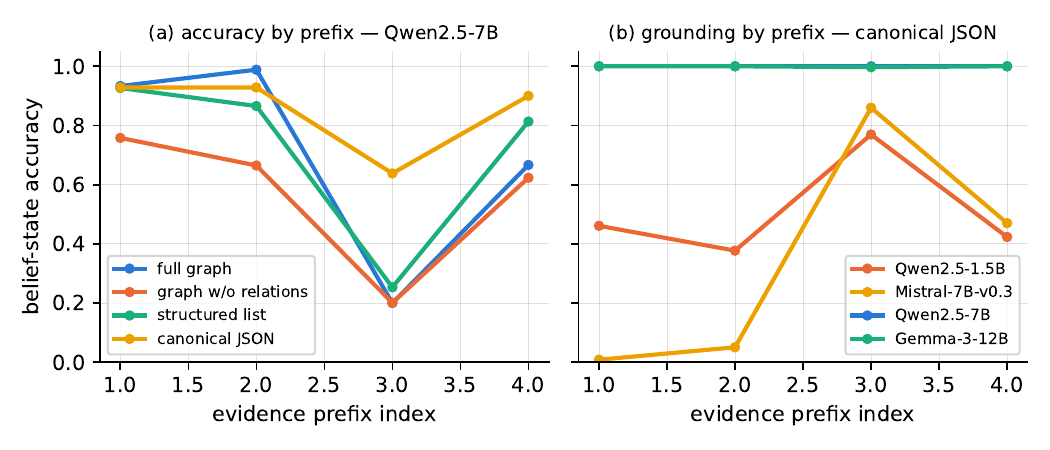}
\caption{(a) Belief-state accuracy by evidence prefix for four Qwen-7B representation conditions. (b) Grounded-prediction rate by prefix for the four models under canonical JSON. The final prefix alone would hide the intermediate failures in both panels.}
\label{fig:prefix-curves}
\end{figure*}

Belief-state accuracy is strongly non-monotonic along the evidence sequence (Fig.~\ref{fig:prefix-curves}a). At the third evidence prefix---where most regimes introduce the report that tests the belief, such as the weak counterevidence that must not reverse it or the opposing claim that opens a contradiction (Fig.~\ref{fig:worked-example})---accuracy for the full-graph, relation-free-graph, and structured-list conditions falls to between 0.20 and 0.25, then recovers to between 0.62 and 0.81 at the fourth prefix; canonical JSON reaches its minimum of 0.64 at the same prefix. Grounding shows the complementary pattern (Fig.~\ref{fig:prefix-curves}b): Mistral's grounded-prediction rate collapses on the first two prefixes and only partially recovers later.

A final-prefix metric would classify most of these trajectories as successful. It would credit the graph-family conditions with 0.62--0.81 accuracy while hiding a prefix at which three quarters of the beliefs were wrong, and it would credit Mistral with its best grounding rate while hiding that it almost never grounds on the initial-evidence situations where a monitoring system has least information. These failures concentrate at the intermediate prefix where the benchmark introduces the evidence that tests preservation, contradiction handling, or revision, rather than being distributed uniformly along the trajectory, and they are shared across serializations. Prefix-level references are what make them measurable.

\subsection{Transparent Baselines and Secondary Diagnostics}
\label{sec:results-baselines}

Among the six fusion baselines (Table~\ref{tab:baselines}), source-reliability weighting scores highest at 0.700 and latest-observation lowest at 0.603; every baseline scores above both diagnostic-tier systems (0.397 and 0.239) and below every core neural system (0.727--0.848). The deterministic reference pipeline's fusion rule attains 0.786 belief accuracy against the normative labels; its action policy selects the normative reference action on 125 of 350 scenarios (35.7\%), matches the reference action family on 78.6\%, and is evidence-supported on 100\%.

\begin{table}[t]
\centering
\footnotesize
\setlength{\tabcolsep}{3.5pt}
\caption{Transparent fusion baselines (main strategies; latest-observation and Dirichlet in the appendix).}
\label{tab:baselines}
\begin{tabular}{lcccc}
\toprule
Fusion baseline & Traj.\ score & Strict acc.\ & Preserv.\ & Revision \\
\midrule
unweighted majority & 0.680 & 0.631 & 0.333 & 0.655 \\
confidence weighted & 0.688 & 0.768 & 0.667 & 0.786 \\
source reliability weighted & 0.700 & 0.856 & 1.000 & 0.856 \\
recency weighted & 0.691 & 0.815 & 0.667 & 0.857 \\
\bottomrule
\end{tabular}

\end{table}

On the endpoint action task, exact reference-policy compliance is 0.214, 0.213, 0.215, and 0.149 for Qwen-1.5B, Mistral-7B, Qwen-7B, and Gemma-12B, while the evidence-supported-action rate is 0.857 for all four. The identical rate is structural: the predicate accepts either \texttt{verify} or \texttt{inspect} in 300 of 350 scenarios and only \texttt{monitor} in the remaining 50, so it cannot distinguish among actions within the investigate family, which is why models with different exact reference-policy compliance (0.149--0.215) receive the same evidence-support score; the metric has limited discriminative value on this suite and is not evidence of equal action reasoning. Contradiction detection accuracy ranges from 0.416 (Qwen-1.5B) to 0.960 (Qwen-7B, no-confidence condition), and Qwen-7B's full-graph localization F1 exceeds 0.95 on the contradiction-resolution and source-reliability regimes against 0.001 for Mistral-7B overall. Strict JSON validity is at or above 0.997 for every core system with repair rates below 0.3\%. Contradiction-type accuracy, failure taxonomies, budget sensitivity, prompt lengths, confidence-bin agreement, the override diagnostic, and qualitative error examples are in the appendix.

\section{Discussion}
\label{sec:discussion}

\subsection{What CREG Reveals About Model Evaluation}
\label{sec:discussion-evaluation}

The panel separates on entity grounding before it separates on belief updating: the diagnostic-tier models emit mostly valid JSON yet lose much of their trajectory score to predictions that never bind to the required entity, and in the exploratory conditional analysis their grounded updates land within a few points of the core models'. We therefore read the model contrasts of Sec.~\ref{sec:results-grounding} as predominantly grounding gaps rather than pure reasoning gaps, subject to the upper-bound caveat. Practically, output-side entity binding should be engineered and verified before poor performance is attributed to fusion or revision competence.

The prefix curves extend this message from grounding to belief. A mid-sequence collapse shared by every graph-family condition, an early-prefix grounding collapse in one model, and the difference between prompt and delayed revision are invisible to endpoint evaluation by construction, because a final answer aggregates over the dynamics one needs to inspect. Endpoint benchmarks remain useful for what they measure; they cannot show whether a system reached its answer through a defensible sequence of beliefs. We regard the prefix curves, rather than any scalar, as the primary scientific output of the trajectory track.

\subsection{What the Representation Study Supports}
\label{sec:discussion-representation}

Relation semantics matter; graph form by itself does not. The relation effect is categorical: large and Holm-significant on strict belief-state accuracy in the regime built to require relations, small and not significant after correction on the regime-scoped trajectory score, so the two metrics register the manipulation differently. Part of the overall deficit persists when relation content is kept in flat form, so content and serialization both contribute; but the information-matched structured list outperforms the full graph and permuting record order slightly improves it, so neither graph serialization nor ordering carries measurable value here. The study manipulates serialization, relation content, record order, and selected metadata fields; it does not test arbitrary graph topologies.

Two features bound this conclusion. The two primary metrics dissociate---canonical JSON yields more exactly correct categorical states while richer serializations track the reference distribution more closely---which we attribute tentatively to shorter prompts favoring decisive commitments, without assigning the difference to format alone given the length confound. And the field ablations resist a clean story: the provenance manipulation is confounded by design and its regime-scoped effects point in opposite directions on the two metrics, the timestamp null is expected because state epochs remain visible, and no dedicated source-reliability ablation exists. These are boundaries on the evidence that the information controls make visible.

\subsection{Action Metrics and Transparent Baselines}
\label{sec:discussion-actions}

All four models agree with the single reference action about a fifth of the time, and the transparent reference pipeline agrees with the normative policy on roughly a third of scenarios while remaining evidence-supported on all of them; exact compliance thus measures agreement with one normative reference policy, not decision quality, and the coarse evidence-support rate (Sec.~\ref{sec:results-baselines}) cannot rank systems within the investigate family. The transparent fusion baselines are informative in both directions: the best inspectable strategy trails the best neural systems by roughly 0.15 trajectory-score points, so simple fusion does not solve the task, while every baseline outscores both diagnostic-tier models. Thus much of the weak-model deficit arises before distributional updating can be evaluated, through ungrounded or malformed outputs; this does not show that their updating is adequate, since the conditional analysis selects a non-random subset of prefixes. That reliability weighting is the best transparent strategy follows from the reference semantics and is not an independent discovery.

\subsection{Limitations and Outlook}
\label{sec:discussion-limitations}

Five limitations bound our claims: the suite is synthetic, with a binary state space and one scored target entity, and the physical-log replay in the appendix demonstrates the ingestion pathway rather than an empirical robot evaluation; the semantic-plausibility audit used one expert and does not establish human consensus; two representation ablations are confounded as disclosed in Sec.~\ref{sec:methods-representations}, the timestamp manipulation is intentionally weak, and the source-reliability factor lacks a dedicated ablation; the conditional trajectory score and Gemma's non-canonical runs are exploratory, and no conclusion rests on them; and the panel spans four open-weight models between 1.5B and 12B parameters at a fixed 512-token budget, so larger models, tool-augmented systems, and longer budgets are untested.

The override diagnostic in the appendix shows that a verifier which always overrides the model discards many evidence-supported proposals (harmful-override rate 0.494), which motivates selective neurosymbolic verification that accepts, abstains, or overrides per decision with calibrated triggers. Richer multi-entity state spaces, longer recorded robotic sequences through the replay pathway, and a dedicated source-reliability manipulation would complete the factor set this benchmark was designed to expose.

\section{Conclusion}
\label{sec:conclusion}

We introduced CREG, a controlled benchmark and layered evaluation protocol for evidence-grounded shared belief updating in Connected Robotics. Across 350 scenarios, four open-weight instruction models, and six transparent fusion baselines, the results show that weak-model performance is strongly affected by entity-grounding failures that occur before belief updating can be judged; that semantic relation content improves categorical belief accuracy where relational evidence is required, while graph serialization alone offers no consistent advantage once information content is controlled; and that final-state evaluation hides substantial transient errors along the belief trajectory.

These findings support evaluating connected-robot reasoning as a layered evidence-to-belief process rather than as a single final decision. By separating grounding, revision, contradiction handling, and action selection, and by exposing a reference belief at every evidence prefix, CREG provides a reproducible basis for studying shared semantic world models, distributed physical reasoning, and the selective decision verification that these results motivate.

\section*{Acknowledgment}
This work was supported by JSPS KAKENHI Grant Number JP26K17338. We gratefully acknowledge the support of NVIDIA Corporation.

\bibliographystyle{IEEEtran}
\bibliography{related_work}

\begin{thebibliography}{10}
\providecommand{\url}[1]{#1}
\csname url@samestyle\endcsname
\providecommand{\newblock}{\relax}
\providecommand{\bibinfo}[2]{#2}
\providecommand{\BIBentrySTDinterwordspacing}{\spaceskip=0pt\relax}
\providecommand{\BIBentryALTinterwordstretchfactor}{4}
\providecommand{\BIBentryALTinterwordspacing}{\spaceskip=\fontdimen2\font plus
\BIBentryALTinterwordstretchfactor\fontdimen3\font minus
  \fontdimen4\font\relax}
\providecommand{\BIBforeignlanguage}[2]{{%
\expandafter\ifx\csname l@#1\endcsname\relax
\typeout{** WARNING: IEEEtran.bst: No hyphenation pattern has been}%
\typeout{** loaded for the language `#1'. Using the pattern for}%
\typeout{** the default language instead.}%
\else
\language=\csname l@#1\endcsname
\fi
#2}}
\providecommand{\BIBdecl}{\relax}
\BIBdecl

\bibitem{Bezerra2026ConnectedRobotics}
R.~Bezerra, S.~Tadokoro, and K.~Ohno, ``{AI--IoT--Robotics} integration: Survey
  of frameworks, emerging trends, and the path toward connected robotics,''
  \emph{IEEE Internet of Things Journal}, vol.~13, no.~10, pp.
  20\,398--20\,412, 2026.

\bibitem{Rosinol2020DynamicSceneGraphs}
\BIBentryALTinterwordspacing
A.~Rosinol, A.~Gupta, M.~Abate, J.~Shi, and L.~Carlone, ``{3D} dynamic scene
  graphs: Actionable spatial perception with places, objects, and humans,'' in
  \emph{Robotics: Science and Systems XVI}.\hskip 1em plus 0.5em minus
  0.4em\relax Robotics: Science and Systems Foundation, 2020. [Online].
  Available: \url{https://www.roboticsproceedings.org/rss16/p079.html}
\BIBentrySTDinterwordspacing

\bibitem{Rosinol2021Kimera}
A.~Rosinol \emph{et~al.}, ``Kimera: From {SLAM} to spatial perception with {3D}
  dynamic scene graphs,'' \emph{The International Journal of Robotics
  Research}, vol.~40, no. 12--14, pp. 1510--1546, 2021.

\bibitem{Hughes2022Hydra}
\BIBentryALTinterwordspacing
N.~Hughes, Y.~Chang, and L.~Carlone, ``Hydra: A real-time spatial perception
  system for {3D} scene graph construction and optimization,'' in
  \emph{Robotics: Science and Systems XVIII}.\hskip 1em plus 0.5em minus
  0.4em\relax New York City, NY, USA: Robotics: Science and Systems Foundation,
  2022. [Online]. Available:
  \url{https://www.roboticsproceedings.org/rss18/p050.html}
\BIBentrySTDinterwordspacing

\bibitem{Chang2023HydraMulti}
Y.~Chang, N.~Hughes, A.~Ray, and L.~Carlone, ``Hydra-multi: Collaborative
  online construction of {3D} scene graphs with multi-robot teams,'' in
  \emph{2023 IEEE/RSJ International Conference on Intelligent Robots and
  Systems (IROS)}.\hskip 1em plus 0.5em minus 0.4em\relax IEEE, 2023, pp.
  10\,995--11\,002.

\bibitem{Gu2024ConceptGraphs}
Q.~Gu \emph{et~al.}, ``Conceptgraphs: Open-vocabulary {3D} scene graphs for
  perception and planning,'' in \emph{2024 IEEE International Conference on
  Robotics and Automation (ICRA)}.\hskip 1em plus 0.5em minus 0.4em\relax IEEE,
  2024, pp. 5021--5028.

\bibitem{Werby2024HOVSG}
\BIBentryALTinterwordspacing
A.~Werby, C.~Huang, M.~B{\"u}chner, A.~Valada, and W.~Burgard, ``Hierarchical
  open-vocabulary {3D} scene graphs for language-grounded robot navigation,''
  in \emph{Robotics: Science and Systems XX}.\hskip 1em plus 0.5em minus
  0.4em\relax Delft, Netherlands: Robotics: Science and Systems Foundation,
  2024. [Online]. Available:
  \url{https://www.roboticsproceedings.org/rss20/p077.html}
\BIBentrySTDinterwordspacing

\bibitem{Ginting2023SemanticBeliefGraph}
M.~F. Ginting \emph{et~al.}, ``Safe and efficient navigation in extreme
  environments using semantic belief graphs,'' in \emph{2023 IEEE International
  Conference on Robotics and Automation (ICRA)}.\hskip 1em plus 0.5em minus
  0.4em\relax IEEE, 2023, pp. 5653--5658.

\bibitem{Ginting2024SemanticBeliefBehaviorGraph}
M.~F. Ginting, D.~D. Fan, S.-K. Kim, M.~J. Kochenderfer, and A.-A.
  Agha-Mohammadi, ``Semantic belief behavior graph: Enabling autonomous robot
  inspection in unknown environments,'' in \emph{2024 IEEE/RSJ International
  Conference on Intelligent Robots and Systems (IROS)}.\hskip 1em plus 0.5em
  minus 0.4em\relax IEEE, 2024, pp. 7604--7610.

\bibitem{Kojima2022HeterogeneousRobots}
S.~Kojima \emph{et~al.}, ``Heterogeneous robots coordination for industrial
  plant inspection and evaluation at world robot summit 2020,'' \emph{Advanced
  Robotics}, vol.~36, no.~21, pp. 1102--1119, 2022.

\bibitem{Dempster1967UpperLower}
A.~P. Dempster, ``Upper and lower probabilities induced by a multivalued
  mapping,'' \emph{The Annals of Mathematical Statistics}, vol.~38, no.~2, pp.
  325--339, 1967.

\bibitem{Julier1997CovarianceIntersection}
S.~J. Julier and J.~K. Uhlmann, ``A non-divergent estimation algorithm in the
  presence of unknown correlations,'' in \emph{Proceedings of the 1997 American
  Control Conference}, vol.~4.\hskip 1em plus 0.5em minus 0.4em\relax IEEE,
  1997, pp. 2369--2373.

\bibitem{Dong2009SourceDependence}
X.~L. Dong, L.~Berti-{\'E}quille, and D.~Srivastava, ``Integrating conflicting
  data: The role of source dependence,'' \emph{Proceedings of the VLDB
  Endowment}, vol.~2, no.~1, pp. 550--561, 2009.

\bibitem{Li2016TruthDiscovery}
Y.~Li \emph{et~al.}, ``A survey on truth discovery,'' \emph{ACM SIGKDD
  Explorations Newsletter}, vol.~17, no.~2, pp. 1--16, 2016.

\bibitem{Alchourron1985TheoryChange}
C.~E. Alchourr{\'o}n, P.~G{\"a}rdenfors, and D.~Makinson, ``On the logic of
  theory change: Partial meet contraction and revision functions,'' \emph{The
  Journal of Symbolic Logic}, vol.~50, no.~2, pp. 510--530, 1985.

\bibitem{Darwiche1997IteratedBeliefRevision}
A.~Darwiche and J.~Pearl, ``On the logic of iterated belief revision,''
  \emph{Artificial Intelligence}, vol.~89, no. 1--2, pp. 1--29, 1997.

\bibitem{Wilie2024BeliefRevision}
\BIBentryALTinterwordspacing
B.~Wilie, S.~Cahyawijaya, E.~Ishii, J.~He, and P.~Fung, ``Belief revision: The
  adaptability of large language models reasoning,'' in \emph{Proceedings of
  the 2024 Conference on Empirical Methods in Natural Language
  Processing}.\hskip 1em plus 0.5em minus 0.4em\relax Miami, Florida, USA:
  Association for Computational Linguistics, 2024, pp. 10\,480--10\,496.
  [Online]. Available: \url{https://aclanthology.org/2024.emnlp-main.586/}
\BIBentrySTDinterwordspacing

\bibitem{Xu2026BeliefTrack}
\BIBentryALTinterwordspacing
H.~Xu \emph{et~al.}, ``When should models change their minds? contextual belief
  management in large language models,'' 2026, arXiv preprint; accepted to
  EMNLP 2026 (main conference) per the arXiv listing. [Online]. Available:
  \url{https://arxiv.org/abs/2605.30219}
\BIBentrySTDinterwordspacing

\bibitem{Samanta2026BayesBench}
\BIBentryALTinterwordspacing
A.~Samanta \emph{et~al.}, ``{BayesBench}: Evaluating {LLM} belief trajectories
  under multi-turn evidence accumulation,'' 2026, preprint. [Online].
  Available: \url{https://arxiv.org/abs/2606.30850}
\BIBentrySTDinterwordspacing

\bibitem{Shridhar2020ALFRED}
M.~Shridhar \emph{et~al.}, ``{ALFRED}: A benchmark for interpreting grounded
  instructions for everyday tasks,'' in \emph{2020 IEEE/CVF Conference on
  Computer Vision and Pattern Recognition (CVPR)}.\hskip 1em plus 0.5em minus
  0.4em\relax IEEE, 2020, pp. 10\,737--10\,746.

\bibitem{Padmakumar2022TEACh}
A.~Padmakumar \emph{et~al.}, ``{TEACh}: Task-driven embodied agents that
  chat,'' \emph{Proceedings of the AAAI Conference on Artificial Intelligence},
  vol.~36, no.~2, pp. 2017--2025, 2022.

\bibitem{Li2023Behavior1K}
\BIBentryALTinterwordspacing
C.~Li \emph{et~al.}, ``{BEHAVIOR-1K}: A benchmark for embodied {AI} with 1,000
  everyday activities and realistic simulation,'' in \emph{Proceedings of the
  6th Conference on Robot Learning}, ser. Proceedings of Machine Learning
  Research, vol. 205.\hskip 1em plus 0.5em minus 0.4em\relax PMLR, 2023, pp.
  80--93. [Online]. Available:
  \url{https://proceedings.mlr.press/v205/li23a.html}
\BIBentrySTDinterwordspacing

\bibitem{Majumdar2024OpenEQA}
A.~Majumdar \emph{et~al.}, ``{OpenEQA}: Embodied question answering in the era
  of foundation models,'' in \emph{2024 IEEE/CVF Conference on Computer Vision
  and Pattern Recognition (CVPR)}.\hskip 1em plus 0.5em minus 0.4em\relax IEEE,
  2024, pp. 16\,488--16\,498.

\bibitem{Ichter2023SayCan}
\BIBentryALTinterwordspacing
B.~Ichter \emph{et~al.}, ``Do as i can, not as i say: Grounding language in
  robotic affordances,'' in \emph{Proceedings of the 6th Conference on Robot
  Learning}, ser. Proceedings of Machine Learning Research, vol. 205.\hskip 1em
  plus 0.5em minus 0.4em\relax PMLR, 2023, pp. 287--318. [Online]. Available:
  \url{https://proceedings.mlr.press/v205/ichter23a.html}
\BIBentrySTDinterwordspacing

\bibitem{Driess2023PaLME}
\BIBentryALTinterwordspacing
D.~Driess \emph{et~al.}, ``{PaLM-E}: An embodied multimodal language model,''
  in \emph{Proceedings of the 40th International Conference on Machine
  Learning}, ser. Proceedings of Machine Learning Research, vol. 202.\hskip 1em
  plus 0.5em minus 0.4em\relax PMLR, 2023, pp. 8469--8488. [Online]. Available:
  \url{https://proceedings.mlr.press/v202/driess23a.html}
\BIBentrySTDinterwordspacing

\bibitem{Zitkovich2023RT2}
\BIBentryALTinterwordspacing
B.~Zitkovich \emph{et~al.}, ``{RT-2}: Vision-language-action models transfer
  web knowledge to robotic control,'' in \emph{Proceedings of the 7th
  Conference on Robot Learning}, ser. Proceedings of Machine Learning Research,
  vol. 229.\hskip 1em plus 0.5em minus 0.4em\relax PMLR, 2023, pp. 2165--2183.
  [Online]. Available:
  \url{https://proceedings.mlr.press/v229/zitkovich23a.html}
\BIBentrySTDinterwordspacing

\bibitem{Rana2023SayPlan}
\BIBentryALTinterwordspacing
K.~Rana, J.~Haviland, S.~Garg, J.~Abou-Chakra, I.~Reid, and N.~Suenderhauf,
  ``Sayplan: Grounding large language models using {3D} scene graphs for
  scalable robot task planning,'' in \emph{Proceedings of the 7th Conference on
  Robot Learning}, ser. Proceedings of Machine Learning Research, vol.
  229.\hskip 1em plus 0.5em minus 0.4em\relax PMLR, 2023, pp. 23--72. [Online].
  Available: \url{https://proceedings.mlr.press/v229/rana23a.html}
\BIBentrySTDinterwordspacing

\bibitem{Sun2024ThinkOnGraph}
\BIBentryALTinterwordspacing
J.~Sun \emph{et~al.}, ``Think-on-graph: Deep and responsible reasoning of large
  language model on knowledge graph,'' in \emph{The Twelfth International
  Conference on Learning Representations}, 2024. [Online]. Available:
  \url{https://openreview.net/forum?id=nnVO1PvbTv}
\BIBentrySTDinterwordspacing

\bibitem{He2024GRetriever}
\BIBentryALTinterwordspacing
X.~He \emph{et~al.}, ``{G-Retriever}: Retrieval-augmented generation for
  textual graph understanding and question answering,'' in \emph{Advances in
  Neural Information Processing Systems}, vol.~37.\hskip 1em plus 0.5em minus
  0.4em\relax Curran Associates, Inc., 2024, pp. 132\,876--132\,907. [Online].
  Available:
  \url{https://proceedings.neurips.cc/paper_files/paper/2024/hash/efaf1c9726648c8ba363a5c927440529-Abstract-Conference.html}
\BIBentrySTDinterwordspacing

\bibitem{DeYoung2020ERASER}
\BIBentryALTinterwordspacing
J.~DeYoung \emph{et~al.}, ``{ERASER}: A benchmark to evaluate rationalized
  {NLP} models,'' in \emph{Proceedings of the 58th Annual Meeting of the
  Association for Computational Linguistics}.\hskip 1em plus 0.5em minus
  0.4em\relax Online: Association for Computational Linguistics, 2020, pp.
  4443--4458. [Online]. Available:
  \url{https://aclanthology.org/2020.acl-main.408/}
\BIBentrySTDinterwordspacing

\bibitem{Jacovi2020Faithfulness}
\BIBentryALTinterwordspacing
A.~Jacovi and Y.~Goldberg, ``Towards faithfully interpretable {NLP} systems:
  How should we define and evaluate faithfulness?'' in \emph{Proceedings of the
  58th Annual Meeting of the Association for Computational Linguistics}.\hskip
  1em plus 0.5em minus 0.4em\relax Online: Association for Computational
  Linguistics, 2020, pp. 4198--4205. [Online]. Available:
  \url{https://aclanthology.org/2020.acl-main.386/}
\BIBentrySTDinterwordspacing

\end{thebibliography}

\appendices
\renewcommand{\thetable}{A\arabic{table}}
\renewcommand{\thefigure}{A\arabic{figure}}
\renewcommand{\theHtable}{A\arabic{table}}
\renewcommand{\theHfigure}{A\arabic{figure}}
\setcounter{table}{0}
\setcounter{figure}{0}
This appendix collects the operational, reproducibility, and secondary-analysis material that supports the main text. No result reported here changes any conclusion of the main text.

\section{Runtime Protocol and Freeze Record}
\label{sec:supp-runtime}

\subsection{Model Panel}
The preregistered panel comprises \texttt{Qwen/Qwen2.5-1.5B-Instruct}, \texttt{Qwen/Qwen2.5-7B-Instruct}, \texttt{mistralai/Mistral-7B-Instruct-v0.3}, and \texttt{google/gemma-3-12b-it}, which the preregistration listed as optional pending access and which was run in full under canonical JSON. Decoding is fixed for all systems: sampling enabled, temperature 0.2, top-$p$ 0.9. Three generation seeds (42, 43, 44) are treated as repeated generations nested within the same 350 scenarios and are averaged within scenario before any inference. The frozen benchmark manifest records the model identifiers, representation settings, prompt hashes, and per-scenario content hashes for every system; the \texttt{model\_revision} and \texttt{tokenizer\_revision} fields in the manifest are unset, so model snapshots are identified by name rather than by commit hash (see Sec.~\ref{sec:supp-concerns}).

\subsection{Token-Budget Selection}
A seven-scenario runtime pilot (one scenario per regime, seed 42) evaluated the preregistered output-budget candidates $\{256, 384, 512, 768\}$ on the twelve-system runtime-fallback panel. No budget produced context overflow or truncation. The 512-token budget gave the highest strict-validity and pooled entity-grounding rates with a projected full-run cost of 45.8\,h, below the preregistered 48\,h cap (Table~\ref{tab:supp-budget}). Because the full model$\times$representation crossing exceeded the cap, the preregistered fallback applied: every model runs canonical JSON, and Qwen2.5-7B alone runs the complete representation and factor-ablation matrix (twelve systems in total). Mean trajectory score on the pilot varies by less than 0.02 across the tested budgets, so the budget choice does not drive the reported conclusions within the tested range; budgets above 512 tokens were not run at full scale.

\begin{table*}[t]
\centering
\footnotesize
\setlength{\tabcolsep}{3pt}
\caption{Token-budget sensitivity on the seven-scenario pilot (seed 42, twelve-system panel). The 512 choice used the pooled grounding rate; the core-scoped rate is marginally highest at 256.}
\label{tab:supp-budget}
\begin{tabular}{rrrrrrrr}
\toprule
Budget & Strict valid & Repair & Grounding (pooled) & Grounding (core) & Truncated & Mean traj.\ & Proj.\ hours \\
\midrule
256 & 0.928 & 0.021 & 0.928 & 0.997 & 0.000 & 0.572 & 39.4 \\
384 & 0.954 & 0.000 & 0.912 & 0.990 & 0.000 & 0.573 & 43.6 \\
512 & 0.961 & 0.003 & 0.930 & 0.986 & 0.000 & 0.589 & 45.8 \\
\bottomrule
\end{tabular}

\end{table*}

\subsection{Diagnostic Tier}
The preregistered runtime gate required a pooled required-entity grounding rate at or above a fixed threshold. The pilot showed that two systems---Qwen2.5-1.5B and Mistral-7B under canonical JSON---would fail that gate for reasons that are a model-capability result rather than an infrastructure problem. A preregistration amendment (\texttt{evaluation-v1.0-draft2}, 2026-07-27, adopted before the freeze) therefore tiered the gate: it is evaluated over \emph{core} systems only, and the two systems are retained in the matrix as a \emph{diagnostic} tier whose grounding failures are reported in the main text rather than pooled into runtime-compatibility acceptance. No scenario, reference, prompt, or metric changed under the amendment.

\subsection{Freeze Gates}
The benchmark was frozen under the annotated tag \texttt{evaluation-v1.0} after seven gates passed: 350 unique scenario identifiers and semantic fingerprints; agreement of the independent log-odds reference checker on every prefix; verification that all relation-association counterfactual pairs are observationally indistinguishable without relations; completion of the 35-case expert audit without unresolved corrections; the runtime pilot meeting all thresholds; the full repository test suite; and the task output-contract tests. Post-freeze corrections are restricted to the report layer (metric aggregation and comparison code) and are logged in the repository changelog; no frozen artifact was modified. Gemma runs on the structured-list, graph-without-relations, and full-graph representations were collected outside the preregistered fallback panel and are reported only in this document, as exploratory results.

\section{Representation Conditions: Additional Detail}
\label{sec:supp-representations}

\begin{figure*}[t]
\centering
\includegraphics[width=0.92\textwidth]{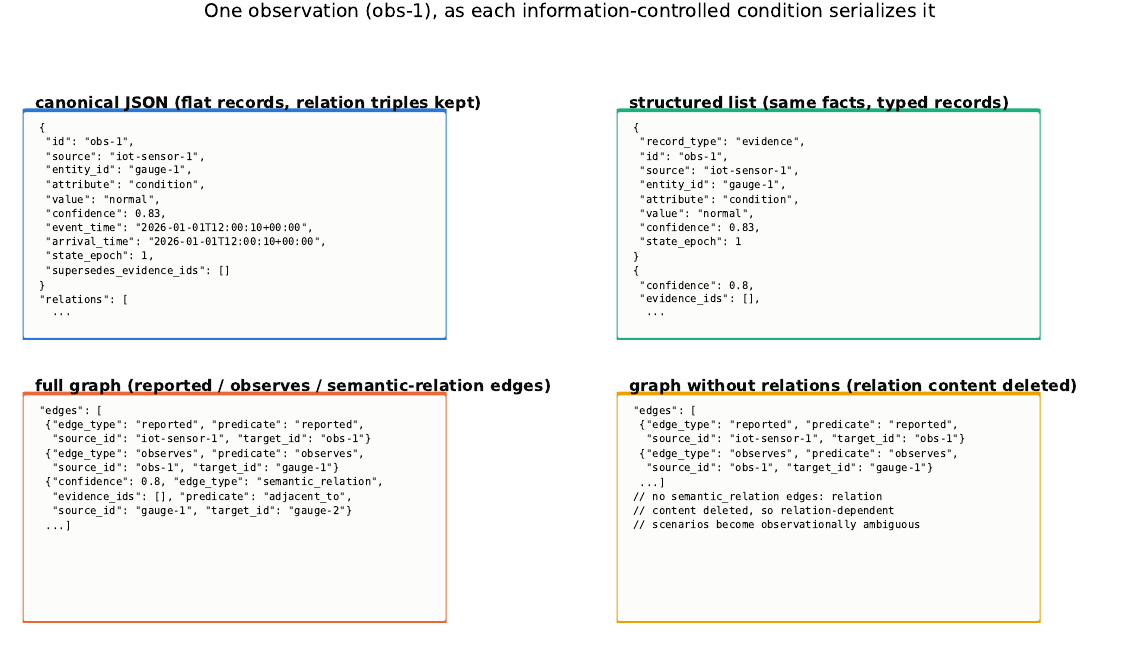}
\caption{One observation from the worked-example scenario of the main text (Fig.~\ref{fig:worked-example}), excerpted verbatim from the frozen prompts of the four principal representation conditions.}
\label{fig:supp-representations}
\end{figure*}

Fig.~\ref{fig:supp-representations} shows the same observation record exactly as the four principal conditions serialize it, excerpted verbatim from the frozen prompts. Table~\ref{tab:supp-lengths} quantifies prompt length for every condition.

\begin{table}[t]
\centering
\footnotesize
\caption{Trajectory-task prompt lengths by condition (seed 42; prompts are seed-invariant).}
\label{tab:supp-lengths}
\resizebox{\linewidth}{!}{
\begin{tabular}{lrrr}
\toprule
Condition & Mean chars & Max chars & Rel.\ to full graph \\
\midrule
canonical\_json & 5284 & 6090 & 0.843 \\
structured\_list & 5651 & 6513 & 0.901 \\
flat-edge-removed-relations & 6298 & 7657 & 1.004 \\
graph\_without\_relations & 6058 & 7430 & 0.966 \\
graph & 6271 & 7624 & 1.000 \\
no-provenance & 5834 & 6821 & 0.930 \\
no-timestamps & 6053 & 7224 & 0.965 \\
no-confidence & 6210 & 7511 & 0.990 \\
shuffled-full-graph & 6271 & 7624 & 1.000 \\
\bottomrule
\end{tabular}
}
\end{table}

Beyond the two confounds stated in the main text (canonical JSON is $0.843\times$ the full-graph length; the no-provenance condition also removes the source-reliability field), the following caveats apply to the factor ablations. The \emph{no-timestamps} condition removes event and arrival times but retains explicit state epochs, so temporal contradiction and supersession semantics remain recoverable; the manipulation is intentionally weak and a null result on the stale/delayed regime is the expected outcome. No \emph{no-source-reliability} ablation exists; the source-reliability regime, which is constructed so that reliability changes the posterior relative to confidence-only fusion, is the available probe for that factor, and the corresponding preregistered hypothesis is only partially testable. The \emph{shuffled full graph} uses a fixed shuffle seed (1729) and is character-identical to the full graph up to record order. The \emph{flat edge-removed} condition preserves relation content as flat records and removes only the graph edge serialization; its length ratio to the full graph is 1.004.

\section{Complete Preregistered Contrast Family}
\label{sec:supp-contrasts}

Table~\ref{tab:supp-contrasts} lists every preregistered paired contrast, including the overall (non-regime-scoped) metadata-field contrasts omitted from the main text. Table~\ref{tab:supp-composition} gives the regime composition of the suite.

\begin{table*}[t]
\centering
\footnotesize
\setlength{\tabcolsep}{5pt}
\caption{All preregistered paired contrasts (trajectory score unless marked; regime-scoped rows restrict to the named regime). $\Delta$ is the mean paired scenario-level difference (system A $-$ system B); $p$ from two-sided sign-flip permutation tests; Holm correction within each family. The Gemma contrast is exploratory (CI only).}
\label{tab:supp-contrasts}
\begin{tabular}{llrrrr}
\toprule
Contrast & Metric & $\Delta$ & 95\% CI & $p$ & Holm $p$ \\
\midrule
graph\_vs\_graph\_without\_relations & trajectory\_score & +0.088 & [+0.072, +0.103] & 0.0000 & 0.0001 \\
graph\_vs\_graph\_without\_relations @ relation\_association & trajectory\_score & +0.020 & [+0.002, +0.037] & 0.0288 & 0.0577 \\
graph\_vs\_edge\_removed & trajectory\_score & +0.041 & [+0.028, +0.053] & 0.0000 & 0.0001 \\
graph\_vs\_canonical\_json & trajectory\_score & +0.017 & [+0.010, +0.024] & 0.0000 & 0.0001 \\
structured\_list\_vs\_canonical\_json & trajectory\_score & +0.045 & [+0.033, +0.056] & 0.0000 & 0.0001 \\
shuffled\_vs\_graph & trajectory\_score & +0.034 & [+0.025, +0.042] & 0.0000 & 0.0001 \\
graph\_vs\_no\_provenance & trajectory\_score & -0.017 & [-0.023, -0.011] & 0.0000 & 0.0001 \\
graph\_vs\_no\_provenance @ source\_reliability & trajectory\_score & +0.019 & [+0.013, +0.025] & 0.0000 & 0.0001 \\
graph\_vs\_no\_timestamps & trajectory\_score & -0.010 & [-0.015, -0.006] & 0.0000 & 0.0001 \\
graph\_vs\_no\_timestamps @ stale\_delayed\_evidence & trajectory\_score & +0.007 & [+0.000, +0.014] & 0.0494 & 0.0577 \\
graph\_vs\_no\_confidence & trajectory\_score & +0.009 & [+0.005, +0.013] & 0.0000 & 0.0001 \\
graph\_vs\_graph\_without\_relations @ relation\_association & strict\_prefix\_belief\_state\_accuracy & +0.153 & [+0.098, +0.207] & 0.0000 & 0.0001 \\
graph\_vs\_no\_provenance @ source\_reliability & strict\_prefix\_belief\_state\_accuracy & -0.200 & [-0.233, -0.167] & 0.0000 & 0.0001 \\
graph\_vs\_no\_timestamps @ stale\_delayed\_evidence & strict\_prefix\_belief\_state\_accuracy & -0.003 & [-0.033, +0.027] & 0.9149 & 0.9149 \\
qwen7b\_vs\_qwen1.5b\_canonical & trajectory\_score & +0.401 & [+0.382, +0.419] & 0.0000 & 0.0000 \\
qwen7b\_vs\_mistral7b\_canonical & trajectory\_score & +0.559 & [+0.545, +0.572] & 0.0000 & 0.0000 \\
gemma12b\_vs\_qwen7b\_canonical (expl.) & trajectory\_score & +0.004 & [-0.005, +0.013] & -- & -- \\
\bottomrule
\end{tabular}

\end{table*}

\begin{table}[t]
\centering
\footnotesize
\caption{Regime composition of the frozen suite.}
\label{tab:supp-composition}
\resizebox{\linewidth}{!}{
\begin{tabular}{lp{6.2cm}r}
\toprule
Regime & Constraint & Scenarios \\
\midrule
justified revision & applicable evidence reverses the top belief & 50 \\
belief preservation & weak counterevidence does not reverse the top belief & 50 \\
relation association & measures relation is necessary to associate sensor evidence with the target & 50 \\
contradiction onset & opposing claims in one epoch create an active contradiction & 50 \\
contradiction resolution & correction evidence supersedes a conflicting claim & 50 \\
stale delayed evidence & older-epoch late evidence is stale while delayed current-epoch evidence applies & 50 \\
source reliability & source reliability changes the posterior relative to confidence-only fusion & 50 \\
\midrule
\multicolumn{2}{l}{Total (3 generation seeds, 512-token budget)} & 350 \\
\bottomrule
\end{tabular}
}
\end{table}

\section{Transparent Baselines and the Deterministic Reference}
\label{sec:supp-baselines}

Six transparent fusion baselines run on the identical suite. Table~\ref{tab:baselines} reports the four main strategies; the two remaining strategies score 0.603 (latest-observation) and 0.678 (Dirichlet-style updater) on trajectory score. Every fusion baseline scores above both diagnostic-tier neural systems and below every core neural system.

The deterministic rule pipeline (graph construction, contradiction detection, confidence-sum fusion, rule-based action selection) is reported as a transparent reference implementation. Its fusion rule attains 0.786 belief accuracy against the normative labels. Its action policy selects the normative reference action on 125 of 350 scenarios (35.7\%), matches the reference action family on 78.6\%, and selects an evidence-supported action on 100\%. The 35.7\% figure measures disagreement between two different policies by design and is not a competence score.

A post-hoc validation configuration in which the deterministic policy overrides model proposals (from Qwen2.5-7B under canonical JSON) is retained as a secondary diagnostic (Table~\ref{tab:supp-hybrid}). Its final action never depends on the proposal, so it is override accounting rather than selective propose-then-verify. The harmful-override rate of 0.494 indicates that an always-override verifier discards many evidence-supported proposals; selective accept/abstain/override verification is left to subsequent work.

\begin{table}[t]
\centering
\footnotesize
\caption{Deterministic-policy override accounting (secondary diagnostic; mean $\pm$ std over three seeds). Final policy accuracy equals the deterministic reference row by construction.}
\label{tab:supp-hybrid}
\resizebox{\linewidth}{!}{
\begin{tabular}{lr}
\toprule
Diagnostic & Mean $\pm$ std (3 seeds) \\
\midrule
Final policy accuracy (= deterministic row; validated action) & 0.357 $\pm$ 0.000 \\
Override rate & 0.637 $\pm$ 0.003 \\
Correct-override rate (unjustified proposal corrected) & 0.143 $\pm$ 0.000 \\
Harmful-override rate (defensible proposal overridden) & 0.494 $\pm$ 0.003 \\
Proposals accepted without correction & 0.363 $\pm$ 0.003 \\
Explanation coverage (model-side) & 0.354 $\pm$ 0.016 \\
Latency per decision (s) & 1.780 $\pm$ 0.001 \\
\bottomrule
\end{tabular}
}
\end{table}

\section{Secondary Diagnostics}
\label{sec:supp-diagnostics}

\subsection{Contradiction Handling, Actions, and Reliability}
On the trajectory track, contradiction detection accuracy ranges from 0.416 (Qwen2.5-1.5B) to 0.960 (Qwen2.5-7B, no-confidence condition). Localization F1 for the Qwen2.5-7B full graph is 0.953 on contradiction-resolution scenarios and 0.993 on source-reliability scenarios, against 0.001 for Mistral-7B overall. Contradiction-type accuracy is sparse because explicit negatives receive no empty-set credit and few systems emit a type label consistently. On the endpoint action task (canonical-JSON systems), exact reference-policy compliance is 0.214 (Qwen2.5-1.5B), 0.213 (Mistral-7B), 0.215 (Qwen2.5-7B), and 0.149 (Gemma-3-12B); the evidence-supported-action rate is 0.857 for all four. Strict JSON validity is at or above 0.997 for every core system with repair rates below 0.3\%; the diagnostic-tier systems reach 0.807 and 0.897.

\subsection{Confidence-Bin Agreement}
Table~\ref{tab:supp-bins} reports ordinal agreement between predicted and reference top-state probability bins over graded (non-tied), grounded prefixes. This is agreement with the normative reference, not empirical calibration. Under canonical JSON, Gemma-3-12B's bin agreement is 0.787 versus 0.905 for Qwen2.5-7B.

\begin{table*}[t]
\centering
\footnotesize
\setlength{\tabcolsep}{3pt}
\caption{Ordinal confidence-bin agreement between predicted and reference top-state probabilities (grounded, non-tied prefixes). Gemma non-canonical rows are exploratory.}
\label{tab:supp-bins}
\begin{tabular}{lrrrr}
\toprule
System & Prefixes & Bin agreement & Pred.\ high-bin & Ref.\ high-bin \\
\midrule
gemma-3-12b-it-canonical\_json & 3297 & 0.787 & 0.730 & 0.711 \\
gemma-3-12b-it-graph & 3276 & 0.885 & 0.764 & 0.693 \\
gemma-3-12b-it-graph\_without\_relations & 3249 & 0.864 & 0.758 & 0.692 \\
gemma-3-12b-it-structured\_list & 3271 & 0.874 & 0.688 & 0.654 \\
mistral-7b-v0.3-canonical\_json & 1054 & 0.710 & 0.310 & 0.226 \\
qwen2.5-1.5b-canonical\_json & 1727 & 0.768 & 0.660 & 0.540 \\
qwen2.5-7b-canonical\_json & 3300 & 0.905 & 0.907 & 0.856 \\
qwen2.5-7b-flat-edge-removed-relations & 3012 & 0.995 & 0.600 & 0.600 \\
qwen2.5-7b-graph & 3300 & 0.960 & 0.706 & 0.700 \\
qwen2.5-7b-graph\_without\_relations & 2844 & 0.998 & 0.581 & 0.581 \\
qwen2.5-7b-no-confidence & 3300 & 0.999 & 0.680 & 0.680 \\
qwen2.5-7b-no-provenance & 3300 & 0.976 & 0.674 & 0.672 \\
qwen2.5-7b-no-timestamps & 3288 & 0.965 & 0.644 & 0.643 \\
qwen2.5-7b-shuffled-full-graph & 3297 & 0.919 & 0.712 & 0.676 \\
qwen2.5-7b-structured\_list & 3182 & 0.990 & 0.634 & 0.634 \\
\bottomrule
\end{tabular}

\end{table*}

\subsection{Failure Taxonomy}
Table~\ref{tab:supp-taxonomy} counts scored failures by category for every system, pooled over regimes and generation seeds. Categories are premature revision, failure to revise, stale-evidence dominance, unreliable-source dominance, contradiction omission, incorrect attribution, relation failure, malformed output, and miscalibration (high Brier distance).

\begin{table*}[t]
\centering
\footnotesize
\setlength{\tabcolsep}{4pt}
\caption{Failure counts by system and category, pooled over regimes and the three generation seeds. Gemma non-canonical rows are exploratory.}
\label{tab:supp-taxonomy}
\begin{tabular}{lrrrrrrrrr}
\toprule
System & prem.\ rev. & fail.\ rev. & stale dom. & unrel.\ dom. & contr.\ om. & inc.\ attr. & rel.\ fail. & malformed & miscal. \\
\midrule
gemma-3-12b-it-canonical\_json & 673 & 252 & 148 & 1 & 292 & 551 & 223 & 3 & 141 \\
gemma-3-12b-it-graph & 523 & 168 & 37 & 17 & 300 & 671 & 258 & 24 & 286 \\
gemma-3-12b-it-graph\_without\_relations & 551 & 157 & 59 & 25 & 302 & 684 & 229 & 51 & 295 \\
gemma-3-12b-it-structured\_list & 530 & 186 & 58 & 7 & 391 & 687 & 280 & 29 & 210 \\
mistral-7b-v0.3-canonical\_json & 1549 & 1153 & 150 & 317 & 764 & 640 & 394 & 2344 & 2536 \\
qwen2.5-1.5b-canonical\_json & 1340 & 950 & 130 & 250 & 994 & 848 & 291 & 1635 & 1868 \\
qwen2.5-7b-canonical\_json & 555 & 5 & 30 & 5 & 803 & 946 & 225 & 0 & 75 \\
qwen2.5-7b-flat-edge-removed-relations & 859 & 468 & 10 & 19 & 266 & 714 & 326 & 288 & 288 \\
qwen2.5-7b-graph & 557 & 465 & 100 & 147 & 328 & 738 & 232 & 0 & 0 \\
qwen2.5-7b-graph\_without\_relations & 914 & 644 & 113 & 103 & 233 & 866 & 301 & 456 & 456 \\
qwen2.5-7b-no-confidence & 485 & 440 & 0 & 140 & 163 & 537 & 252 & 0 & 1 \\
qwen2.5-7b-no-provenance & 420 & 469 & 5 & 57 & 321 & 774 & 228 & 0 & 0 \\
qwen2.5-7b-no-timestamps & 558 & 512 & 98 & 146 & 271 & 708 & 235 & 12 & 12 \\
qwen2.5-7b-shuffled-full-graph & 550 & 316 & 2 & 137 & 466 & 1074 & 265 & 3 & 29 \\
qwen2.5-7b-structured\_list & 583 & 475 & 1 & 44 & 160 & 627 & 268 & 118 & 118 \\
\bottomrule
\end{tabular}

\end{table*}

\section{Expert Audit Protocol}
\label{sec:supp-audit}

One independent expert reviewed a deterministic, stratified sample of 35 scenario--prefix pairs (five per regime). For each case the reviewer recorded, without access to the generated references, the expected top belief, probability direction, supporting- and counterevidence sets, and contradiction status; locked those answers; and only then compared them with the oracle. Round one accepted 29 of 35 cases. The six disagreements, in the relation-association, source-reliability, and stale/delayed regimes, traced to one specification-level issue (tied-posterior semantics), which was repaired in the normative specification; the suite was regenerated and round two accepted all 35 cases with no unresolved issues. No label was patched manually at any point. The audit is a semantic-plausibility sanity check on the generator and specification; it does not establish human consensus, human belief-revision behavior, or model--human agreement, and no inter-rater statistics are reported.

\section{Physical-Log Replay Demonstration}
\label{sec:supp-replay}

The toolkit includes an offline pathway from recorded observation logs to the same evidence-graph, contradiction, belief, decision, and explanation outputs used in the benchmark. The repository's sample testbed (\texttt{distributed\_gauge\_inspection}) contains a two-observation recorded log; the replay therefore demonstrates that the pathway works end to end and is not an empirical robot evaluation. In stepwise replay, the first observation (robot-1 reports gauge-1 \texttt{normal}, confidence 0.82) yields no contradiction and a \texttt{monitor} decision; the second (robot-1 reports \texttt{abnormal}, confidence 0.77) raises a contradiction between the two observations (severity 0.631) and escalates the decision to \texttt{verify}. The final fused belief is normal 0.516 / abnormal 0.484, a near-tie that reflects two conflicting, similarly confident observations. All metadata fields carried by the physical schema (pose, frame identifier, image path, modality, run and sequence identifiers) survive conversion and appear in the contradiction records, so replayed logs retain full provenance for audit.

\section{Qualitative Error Examples}
\label{sec:supp-errors}

The following excerpts (generation seed 42) illustrate the main failure categories. Outputs are truncated.

\paragraph{Ungrounded missing prediction (Mistral-7B, canonical JSON, justified-revision scenario, prefix 1)} The model wraps its answer in prose (``Based on the provided structured semantic evidence, the shared belief after this arrival step can be represented as follows: \ldots'') and emits a distribution over a single state (\texttt{"normal": 0.76}) without the required entity binding; the output is parsed but scores as no prediction for the target entity.

\paragraph{Premature revision (Qwen2.5-7B, canonical JSON, belief-preservation scenario, prefix 3)} On the arrival of weak counterevidence the model shifts the top state to \texttt{anomalous} with probability 0.53, citing only the new report as supporting evidence; the reference preserves the prior top state.

\paragraph{Stale-evidence dominance (Qwen2.5-7B, no-timestamps condition, stale/delayed scenario, prefix 4)} With event and arrival times removed, the model treats an older-epoch late arrival as current, returns a tied distribution (0.5/0.5, \texttt{undetermined}), and lists both the stale report and the current one as supporting evidence; the reference applies only the current-epoch report.

\section{Remaining Reproducibility Notes}
\label{sec:supp-concerns}

Two items are recorded for completeness. First, the frozen manifest identifies models by repository name; the revision-hash fields are unset, so exact reproduction depends on the model snapshots available under those names at run time. Second, per-scenario raw model outputs and the statistics workspace are distributed as release artifacts rather than in the source repository; all tables and figures in the main text are regenerated from those artifacts by the released analysis scripts without re-running any model.

\end{document}